\documentclass{article}

    \PassOptionsToPackage{numbers, compress}{natbib}
\usepackage{iclr2027_conference,times}

\usepackage[utf8]{inputenc} 
\usepackage[T1]{fontenc}    
\usepackage{url}            
\usepackage{booktabs}       
\usepackage{amsfonts}       
\usepackage{nicefrac}       
\usepackage{microtype}      

\usepackage{amsmath,amssymb,amsthm}
\usepackage{mathtools}

\usepackage{booktabs}

\usepackage{tikz}
\usetikzlibrary{arrows.meta,positioning,shapes,calc,fit,backgrounds}

\usepackage{graphicx}
\graphicspath{{figures-images/}}

\usepackage[hidelinks]{hyperref}

\usepackage{multirow}
\usepackage{pifont}
\usepackage{tabularx}
\usepackage{wrapfig}
\usepackage{xcolor}
\usepackage{enumitem}

\usepackage{listings}
\usepackage{xcolor}
\usepackage{makecell}
\definecolor{lightgray}{gray}{0.95}

\lstdefinestyle{promptstyle}{
    backgroundcolor=\color{lightgray},
    basicstyle=\ttfamily\scriptsize,
    breaklines=true,
    breakatwhitespace=true,
    postbreak=\mbox{\textcolor{gray}{$\hookrightarrow$}\space},
    frame=single,
    columns=fullflexible,
    keepspaces=true,
    showstringspaces=false
}

\usepackage{graphicx}
\usepackage{subcaption}

\newcommand{\dataname}{\textsc{ToolDex}}
\definecolor{darkgreen}{rgb}{0.0, 0.5, 0.0}

\usepackage{hyperref}       

\definecolor{linkblue}{rgb}{0.10, 0.40, 0.70}
\hypersetup{
colorlinks=true,
citecolor=linkblue,
linkcolor=linkblue,
urlcolor=linkblue,
}
\iclrfinalcopy

\title{\dataname: An Ego-Exo Human Demonstration Dataset for Creative Physical Tool Use}

\author{%
\normalfont
Mohamed Elmoghany\textsuperscript{1,*},
Kehan Li\textsuperscript{1,*},
Ruoqu Chen\textsuperscript{1,*},
Tarun Chiruvolu\textsuperscript{1},
Shutong Zhang\textsuperscript{1},\\
\normalfont
Ryan Wang\textsuperscript{2},
Stephen Tian\textsuperscript{1},
Yaru Niu\textsuperscript{2},
Michelle Guo\textsuperscript{1},
Ding Zhao\textsuperscript{2},
Weiyu Liu\textsuperscript{1},
Mengdi Xu\textsuperscript{1},\\
\normalfont
Li Fei-Fei\textsuperscript{1},
Jiajun Wu\textsuperscript{1} \\[0.8em]
\normalfont
\textsuperscript{1}Stanford University \quad
\textsuperscript{2}Carnegie Mellon University 
}

\begin{document}

\maketitle
\begingroup
\renewcommand{\thefootnote}{*}
\footnotetext[1]{Equal contribution.}
\endgroup

\fancyhead{}
\renewcommand{\headrulewidth}{0pt}

\begin{abstract}
Creative tool use is a fundamental human capability and an important step toward general-purpose embodied intelligence. It requires agents to reason about object affordances, physical constraints, and manipulation strategies, especially when standard tools are unavailable. However, progress in this area is limited by the lack of real-world datasets that capture human tool-use behavior with sufficient scale, fidelity, and diversity. Existing benchmarks are often simulation-based, focus on simplified interactions, or lack detailed observations and annotations.
We introduce \dataname, a large-scale egocentric-exocentric (ego-exo) human demonstration dataset for creative physical tool use. \dataname~captures scenarios in which humans select, adapt, and repurpose available objects as tools. Using synchronized egocentric cameras and eight exocentric cameras, the dataset provides high-fidelity multi-view observations and rich annotations for hands, tools, target objects, and their deformable interactions. Beyond data collection, we establish a benchmark suite characterizing three core capabilities: physical tool-use reasoning, sparse-view 3D and 4D reconstruction, and robot policy learning. \dataname~provides a foundation for studying human tool-use behavior and developing embodied agents that can adaptively use tools in complex real-world environments. 
Dataset visualizations and benchmarking results are available at \url{https://tooluse-dataset.github.io/}.
\end{abstract}

\section{Introduction}

Tool use is a fundamental human capability and a key step toward general-purpose embodied intelligence~\citep{goodall1964tool, hunt1996manufacture, shumaker2024animal, fitzgerald2021modeling}. In everyday environments, humans do not rely solely on a fixed set of predefined tools; they also flexibly repurpose available objects based on their physical properties and functional affordances. For example, when a knife is unavailable, a person may use a spoon, chopsticks, or another nearby object to divide a cake. This form of creative tool use requires more than recognizing object categories. It demands physical reasoning about contact, geometry and task constraints. As such, physical tool use provides a natural testbed for evaluating whether embodied agents can generalize beyond memorized object-action associations and adapt to novel physical scenarios.

Despite recent progress in robotic manipulation \citep{black2024pi_0, intelligence2025pi_05,kim2026cosmos}, learning generalizable tool-use behavior remains difficult \citep{shi2023robocook, wang2024poco}. Task and motion planning methods can solve structured tool-use problems when
symbolic task descriptions are carefully specified, but often struggle in open-ended real-world settings~\citep{garrett2021integrated}. Recent large language models and vision-language models have shown promising reasoning ability for selecting or describing tools, yet they remain limited in grounding such reasoning into physically executable manipulation~\citep{, liu2024moka,xu2023creative,lin2025robotsmith,gao2025vlmgineer}. This gap is especially pronounced in tasks involving deformable target objects (such as sand or cake), non-rigid contacts, 
and long-horizon interactions, 
where success depends on fine-grained hand-object-tool dynamics rather than high-level semantic reasoning alone.

A major obstacle is \emph{the lack of datasets and benchmarks that capture human physical tool-use capabilities at sufficient scale, fidelity, and diversity}.
Existing robot tool-use benchmarks are often simulation-based or focus on simplified object interactions, making it difficult to evaluate the complexity of real-world manipulation~\citep{xu2023creative, gao2025vlmgineer,bakhtin2019phyre, zhang2025phystoolbench}. 
Meanwhile, recent egocentric human manipulation datasets have greatly expanded the coverage of hand-object interactions~\citep{hoque2025egodex,punamiya2026egoverse,Li2026egolive,liu2024taco}.
Among them, TACO explicitly captures tool-action-object compositions for hand-object manipulation~\citep{liu2024taco}. 
However, it does not explicitly model the target objects manipulated by the tools, such as cakes to be cut or water to be stirred. 
As a result, existing datasets either contain limited tool-use instances or lack the full hand--tool--target-object interactions and annotations needed to study physical tool use.

\begin{figure}[t]
    \centering
    \includegraphics[width=\textwidth]{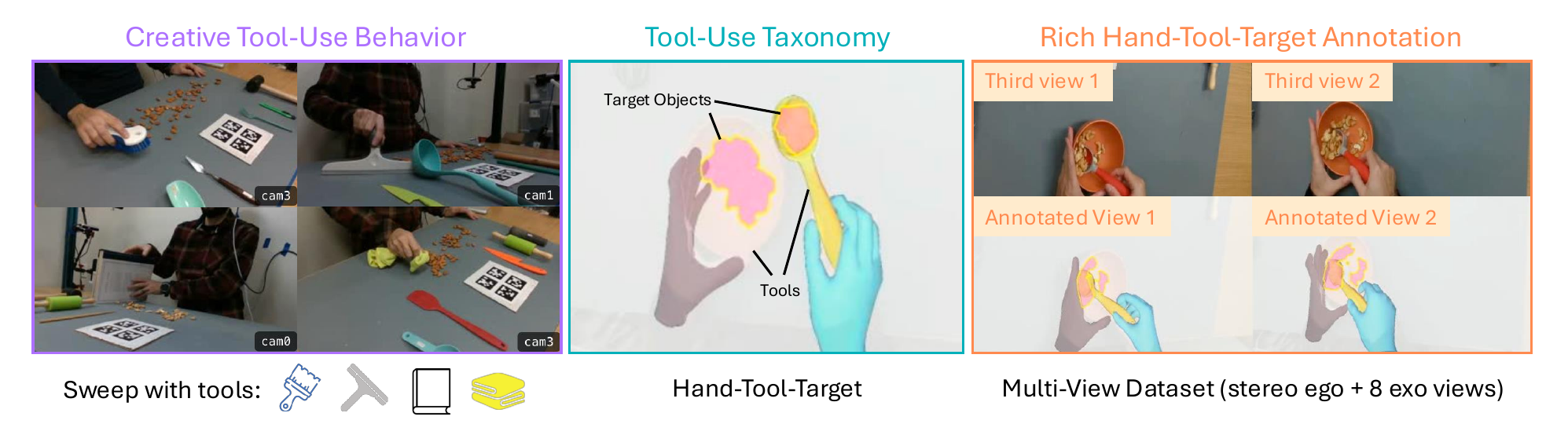}
    \caption{\dataname~Overview.
    The dataset captures \emph{creative physical tool-use behaviors} in everyday tabletop scenarios, where humans repurpose available objects as tools for manipulation (left). 
    It explicitly models full \emph{hand--tool--target-object} interactions, including for challenging deformable targets (middle). 
    Using synchronized ego-exo multi-view capture, \dataname~provides rich annotations across views, including hand pose, tool, and target-object segmentation (right).
    }
    \label{fig:teaser}
    \vspace{-0.237in}
\end{figure}

In this paper, we introduce \dataname, a large-scale ego-exo human demonstration dataset for learning and benchmarking creative tool-based manipulation (\autoref{fig:teaser}). 
Our dataset is designed around everyday tool-use scenarios in which humans \emph{creatively solve manipulation tasks} by selecting, adapting, and repurposing available objects.
A key feature of our dataset is its \emph{explicit coverage of deformable target objects}, such as dough, fluids, and sand, where success depends on fine-grained contact and material interactions. 
To capture these behaviors with high fidelity, we use a synchronized multi-view capture system consisting of two egocentric cameras and eight exocentric cameras. Egocentric views capture local interaction context, while exocentric views enable accurate multi-view tracking and reconstruction.
The resulting dataset provides rich annotations for studying both high-level tool-use reasoning and low-level physical manipulation.

Beyond data collection, we use \dataname~to establish a benchmark suite for evaluating tool-use capabilities across three levels. 
First, we evaluate \textbf{physical tool-use reasoning}, testing whether vision-language models can identify appropriate tools, infer object positions, and provide plausible manipulation strategies.
Second, we evaluate \textbf{3D and 4D reconstruction}, focusing on the recovering hand, tool, and deformable scene states from multi-view observations. 
Third, we evaluate \textbf{robot policy learning}, where human demonstrations are used to train robot policies for tool-based manipulation. 
Together, these benchmarks characterize semantic reasoning, spatial perception, and embodied control, offering a comprehensive evaluation protocol for physical tool-use.

Our contributions are threefold. 
First, we present a real-world ego-exo dataset centered on creative physical tool use spanning diverse tools, deformable target objects, tasks, and interaction strategies. 
Second, we provide rich multi-view annotations that support hand tracking, tool tracking, object interaction analysis, and scene reconstruction. 
Third, we introduce a benchmark suite that evaluates embodied agents from reasoning to reconstruction to policy execution. 
We hope this dataset will serve as a foundation for studying human tool-use behavior and for developing robots that can creatively use tools in complex, open-world environments.

\section{Related Work}

\textbf{Human manipulation datasets.}
Large-scale human video datasets have substantially advanced research on egocentric perception, activity understanding, and human-object interaction~\citep{damen2022rescaling, grauman2022ego4d, grauman2024egoexo4d}. 
A related line of work focuses on hand-object interaction and manipulation~\citep{chao2021dexycb, liu2022hoi4d, yang2022oakink,zhan2024oakink2, banerjee2024introducing, fu2025gigahands, hoque2025egodex, punamiya2026egoverse, Li2026egolive}, shown in \autoref{tab:dataset-comparison}.
HOI4D provides RGB-D observations with 3D hand and object annotations~\citep{liu2022hoi4d}, OakInk focuses on affordance-aware hand-object interactions~\citep{yang2022oakink, zhan2024oakink2}, HOT3D enables accurate multi-view hand-object tracking~\citep{banerjee2024introducing}, and GigaHands scales bimanual manipulation data~\citep{fu2025gigahands}. 
However, they are not designed to systematically capture \emph{creative physical tool use} in everyday scenarios. They focus on general activities, rigid-object interaction, or affordance modeling and lack detailed hand-tool-target-object interactions, and coverage of deformable objects.
\dataname~explicitly targets creative physical tool use and provides data and benchmarks for both high-level reasoning and low-level physical interaction.

\textbf{Tool-use and physical reasoning benchmarks.}
Recent work has increasingly explored tool use and physical reasoning as core aspects of embodied intelligence. 
TACO is closely related to tool-based manipulation, providing a bimanual hand-object interaction dataset with diverse tool--action--object compositions~\citep{liu2024taco}. 
However, it does not model the target objects being acted upon, limiting its ability to capture full tool-use dynamics. 
Beyond datasets, several benchmarks study physical reasoning and tool use from complementary perspectives. PHYRE evaluates agents on physics-based puzzle solving, emphasizing intuitive reasoning ~\citep{bakhtin2019phyre}, while PhysToolBench focuses on tool-use understanding in vision-language models, such as tool selection and strategy explanation~\citep{zhang2025phystoolbench}. 
In robotics, prior work explores creative tool use and planning under physical constraints~\citep{xu2023creative}, as well as leveraging vision-language models for tool design and manipulation~\citep{gao2025vlmgineer}. 
While these efforts highlight the importance of tool use, they either focus on reasoning without real-world interaction data or rely on simplified or simulated settings. 
In contrast, \dataname~provides real-world ego-exo demonstrations and a unified benchmark that connects tool-use reasoning, physical perception, and policy learning.
\section{The \dataname~Dataset}
\label{sec:dataset}

\begin{table}[t]
\centering
\caption{Comparison of \dataname~with recent hand--object-interaction datasets. Only \dataname~explicitly separates \emph{tool} and \emph{target} object roles (Section~\ref{sec:problem}). For prior datasets the dash in \#Target~Obj. indicates a single combined object inventory rather than the absence of targets.}
\label{tab:dataset-comparison}
\setlength{\tabcolsep}{4pt}
\renewcommand{\arraystretch}{1.15}
\resizebox{\textwidth}{!}{%
\begin{tabular}{lcccccccccc}
\toprule
\textbf{Dataset} & \textbf{\#Tools} & \textbf{\#Target Obj.} & \textbf{\#Actions} & \textbf{\#Frames} & \textbf{\#Trials} & \textbf{Ego/Exo} & \textbf{Channels} & \textbf{Hand Pose} & \textbf{Obj. Pose} & \textbf{Seg. Mask} \\
\midrule
TACO~\cite{liu2024taco}            & 20  & --           & 15             & 5.2M  & 2.5K  & 1 / 12   & RGB (Ego RGB-D) & \checkmark & \checkmark & \checkmark \\
OakInk2~\cite{zhan2024oakink2}      & 75  & --           & 60 primitives  & 4.01M & 2.8K  & 1 / 3    & RGB                 & \checkmark & \checkmark & \checkmark \\
HOT3D~\cite{banerjee2024introducing}          & 33  & --           & --             & 3.7M  & 2.8K  & 2--3 / 0 & RGB + mono          & \checkmark & \checkmark & \checkmark \\
GigaHands~\cite{fu2025gigahands}  & 417 & --           & 1467           & 183M  & 13.9K & 0 / 51   & RGB                 & \checkmark & \checkmark & \checkmark \\
\midrule
\textbf{TOOLDEX}     & \textbf{45} & \textbf{23}  & \textbf{16}    & \textbf{58.3M} & \textbf{10.8K} & \textbf{1 / 8} & \textbf{RGB-D}      & \textbf{\checkmark} & \textbf{\checkmark} & \textbf{\checkmark} \\
\bottomrule
\end{tabular}%
}
 \vspace{-0.2in}
\end{table}

\subsection{Problem Formulation}
\label{sec:problem}

\paragraph{Target-centric tool manipulation.}
Tool manipulation, following work on tool use in human and animal cognition~\cite{beck1980,stamant2008,shumaker2011}
, is the indirect alteration of a \emph{target object}'s physical state by an agent acting through a \emph{tool object}, in a scene that may also contain \emph{support objects}, such as containers or work surfaces, that constrain the interaction without being directly actuated. The agent does not act on the target object (e.g., a cake on a plate) directly: it actuates the tool, and the tool's contact with the target produces the state change. We adopt this three-role decomposition---hand, tool, and target object---because it is the \emph{interaction} between tool and target, shaped by tool geometry, target physical properties, and environmental support, that determines whether a manipulation succeeds. \dataname~is designed to capture precisely these interactions.

\paragraph{Tool manipulation as a goal-conditioned MDP.}
We formalize a tool-manipulation trial as a goal-conditioned Markov decision process $(\mathcal{S}, \mathcal{A}, T, \gamma, \mathcal{G})$. The state $s_t \in \mathcal{S}$ is the configuration of all participants of the scene---tool, target, support, and agent---at time $t$. The action $a_t \in \mathcal{A}$ is decomposed into two levels: a \emph{tool action} that determines how the tool moves and, through contact, how the target's state changes; and an \emph{agent action} that determines how the agent's manipulator moves and thereby controls the tool. The transition $T$ captures the physics of the interaction. The goal $g \in \mathcal{G}$ is the desired terminal state of the target object, specified per trial in two complementary ways: a natural-language description $l$, and a terminal-observation $s_T$ from a successful demonstration. A trial is a trajectory $\tau = (s_0, a_0, s_1, a_1, \ldots, a_{T-1}, s_T)$ paired with language goal $l$, and \dataname~is a collection of $(\tau, l)$ pairs.

\subsection{Dataset Properties and Statistics}
\label{sec:analyses}

\begin{figure}[t]
    \centering
    \includegraphics[width=\linewidth]{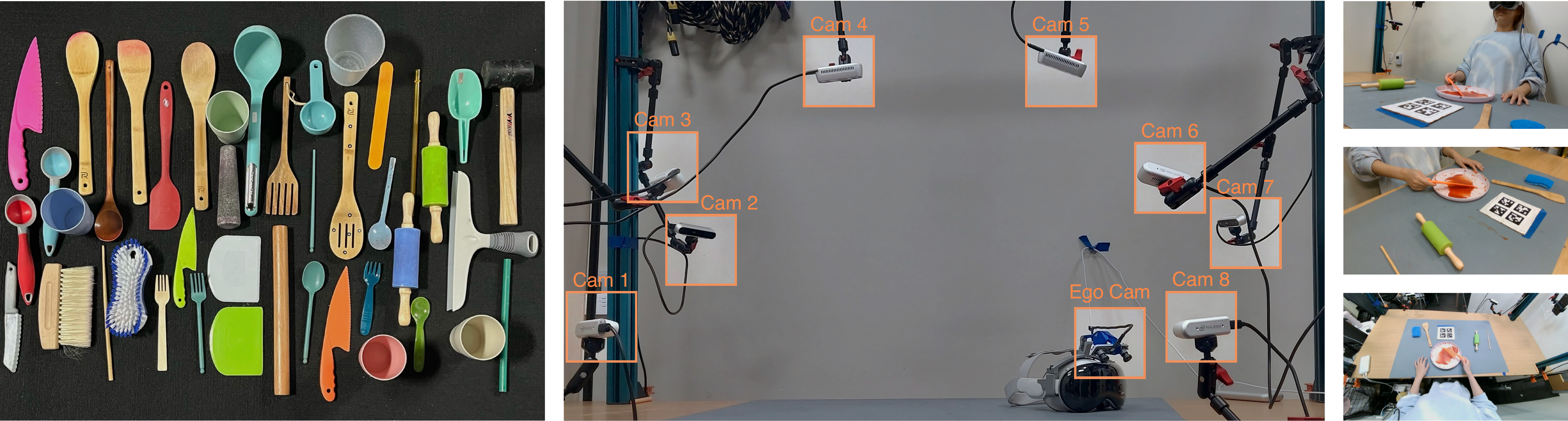}
    \caption{\textbf{Data collection setup.} \emph{Left:} The diverse tool objects in \textsc{ToolDex}. \emph{Middle:} The multi-camera rig, consisting of eight Intel RealSense RGB-D cameras and an Apple Vision Pro headset with a mounted ZED Mini stereo camera. \emph{Right:} Three sampled camera views, including two exocentric and one egocentric view.}
    \label{fig:collection}
     \vspace{-0.2in}
\end{figure}

\paragraph{Diversity of tools, target objects, and actions.}
\dataname~contains 10,853 trials spanning 60 high-level (tool, action, target) task combinations, and 168 total variants that capture changes in tool instances and task attributes (e.g., different tool materials or shapes, and variations in target configurations). In total, the dataset includes 45 tool objects, 23 support objects, 16 deformable object types, and 16 action types. As shown in \autoref{fig:collection}, the tools span a range of geometries, including flat blades (knife, spatula), volumetric scoopers (spoon, ladle, ice-cream scooper, open scooper, measuring scooper), elongated rods (rolling pin, rolling rod, chopstick, straw), and bulkier creative-repurposing items (book, cup, pan), each affording a different contact profile with the target. Target objects span a wide range of physical properties: rigid (toy, box, banana), granular (nuts, sand, kinetic sand, coffee), deformable (dough, naan, towel, tomato sauce). For each (tool, action, target) combination, we collect trials for 60 minutes in which we vary the initial pose of the tool, target, and support, and additionally place 4--5 distractor tools on the work surface to produce realistic clutter and to require visual grounding of the active tool when conditioned on language.

\begin{figure}[t]
    \centering
    \includegraphics[width=\linewidth]{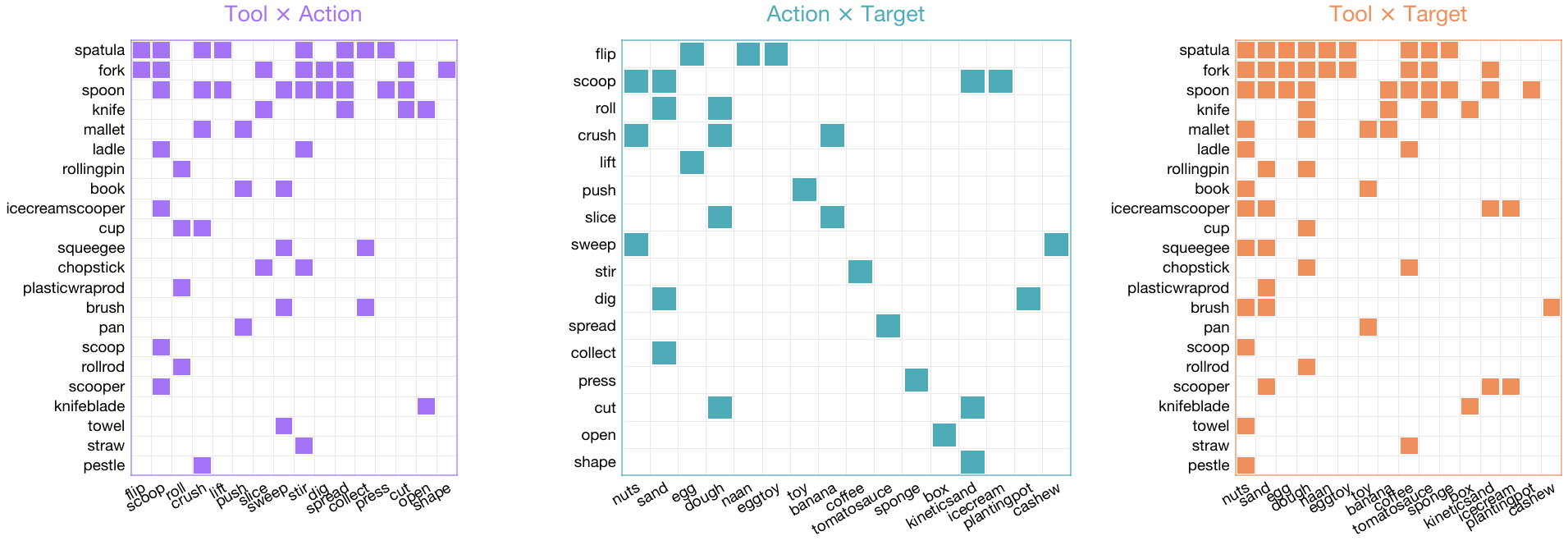}
    \caption{\textbf{Combinatorial coverage of \dataname.} Binary heatmaps of the \emph{(tool, action, target)} combinations in the dataset, with each axis sorted from most to least frequent.}
    \label{fig:combo}
    \vspace{-0.1in}
\end{figure}

\paragraph{Zero-shot and compositional generalization.}
The combinatorial structure of our dataset enables the study of both zero-shot and compositional generalization. As shown in \autoref{fig:combo}, each tool participates in 2.3 actions on average, and each action is realized by 3.6 tools on average across targets. From these (tool, action, target) triples, we construct two families of evaluation splits. \emph{Zero-shot generalization} encompasses three cases, testing the model on a withheld tool, target, or action. \emph{Compositional generalization} holds out a co-occurrence rather than an element, specifically a (tool, action, target) combination in which each individual component has been seen but their joint has not. For example, a held-out triple may evaluate a known tool performing a known action on a target it has not been paired with at training, even though that target has been manipulated by other tools.

\paragraph{Multi-view capture of 3D interaction.}
We capture each interaction with high fidelity to recover the 3D state of the tool, target, and hand, which together provide the inputs required to learn the interaction function T defined in Section \ref{sec:problem}. Each trial records eight calibrated exocentric Intel RealSense RGB-D streams at $1280\times720$ resolution and 30~Hz, together with a stereo egocentric stream from a ZED Mini stereo camera mounted on an Apple Vision Pro headset worn by the demonstrator. Depth observations from each exocentric viewpoint enable 3D reconstruction of state changes, while synchronized multi-view videos capture non-rigid phenomena such as dough deformation and granular flow. 
We provide per-frame multi-view segmentation masks for the \emph{target object}, and tool and hand.  Each trial lasts an average of 18.92 seconds and contains 567.58 frames.

\paragraph{Embodiment-agnostic data representation and collection.}
We collect human data rather than teleoperated robot data because we aim to record \emph{interactions} instead of embodiment-specific motor commands, and because humans achieve substantially higher throughput on creative tool-substitution trials than teleoperation systems. To make the human data useful for robot learning, we record demonstrations at two levels. The \emph{tool action}, represented by the 6-DoF tool pose, captures the physics of how the tool affects the target. This representation is embodiment-invariant and directly transferable to any manipulator capable of reproducing the trajectory. The \emph{agent action}, represented by the MANO hand pose and wrist trajectory, provides the human-side supervision required by retargeting pipelines such as AnyTeleop~\cite{qin2023anyteleop} to map demonstrations onto a robot dexterous hand. In total, \dataname~contains 60 hours of recorded demonstrations across 10,853 trials and 6,158,809 annotated frames, collected at a throughput of approximately 60 trials per hour.

\section{Data Collection}
\label{sec:collection}




Following the task taxonomy introduced in Section~\ref{sec:dataset}, we collect each demonstration in a single calibrated multi-camera studio rig and run an automated annotation pipeline offline after each capture session. The capture rig and annotation pipeline are designed jointly so that each recorded trial yields synchronized RGB-D video, per-frame 6-DoF tool pose, MANO hand poses for both hands, and multi-view segmentation masks for the target object.

\subsection{Task and Object Curation}
Each task combination is a (\emph{tool, action, target}) triplet specifying a set of demonstrations. We curated the task combinations in two stages. We first enumerated 16 everyday tabletop tool-use scenarios drawn from cooking, food preparation, child play, and household activities, each defined by an action verb and a default tool--target pair. We then expanded each scenario along the action$\times$tool axis. For each action, we explicitly enumerated the alternative tools a human might creatively substitute when the canonical tool is unavailable. We select tool, target, and support objects to span diverse geometries, material properties, covering rigid, granular, and deformable objects. Within each task combination, we vary the initial pose of the tool and target object on the work surface, and additionally place 4-5 tool objects as distractor objects in the scene to match the cluttered tabletop conditions of real deployments and require visual grounding of the correct tool when conditioned on language. Each task combination is paired with a natural-language instruction that specifies both the tool motion (e.g., ``swing the spatula laterally under the egg, then lift'') and the intended effect on the target (``flip the egg so its underside faces up'').

\subsection{Hardware Setup}
Our capture rig (\autoref{fig:collection}) consists of 8 Intel RealSense 435 RGB-D cameras and an Apple Vision Pro headset fitted with a Stereolabs ZED Mini stereo camera. The 8 RGB-D cameras are mounted on a Vention extruded-aluminum frame surrounding the work surface, in positions designed to maximize joint coverage of the work volume while preserving sufficient view overlap for multi-view triangulation, 6-DoF object pose estimation, and 4D scene reconstruction. RGB and depth streams are recorded at $1280\times720$ resolution and 30~Hz; streams across the 8 cameras are synchronized in software. We calibrate the cameras using an April Tag followed by multi-view bundle adjustment to obtain rig-relative camera parameters. The ZED Mini, mounted on the Vision Pro, provides two additional $1280\times720$ stereo RGB egocentric streams. To support 6-DoF tool pose estimation, we scan each tool with a Shining3D scanner to obtain a high-resolution textured mesh. The sparse 8 camera setup overcomes occlusion and enables photo-realistic 3D and 4D scene reconstruction.

\subsection{Data Collection and Annotation}
Data is collected under a scripted protocol with a predetermined execution procedure.
We collect one hour of data per task. We also vary the positions of the 4--5 distractor tools across recordings to introduce greater spatial diversity. All 8 exocentric RGB-D streams and the ZED Mini stereo stream are recorded synchronously and stored together. We then run an annotation pipeline that produces three synchronized outputs for each trial: 6-DoF tool pose, MANO hand pose for both hands, and per-frame multi-view segmentation masks for the target object. All annotations are manually verified for each trajectory. When automatic segmentation or tracking fails, annotators provide additional initialization prompts to correct the annotations.

\textbf{Target-object segmentation.} We segment target objects across 8 exocentric views using SAM2~\citep{sam2} with DINOV2~\citep{oquab2023dinov2} seed initialization using the cosine similarity of the scanned tool meshes; SAM2's built-in video tracker propagates the mask across the rest of the trial.

\textbf{Tool pose.} We obtain per-frame 6-DoF tool pose by running FoundationPose~\citep{foundationpose} independently on each exocentric stream, conditioned on the scanned textured mesh of the tool used in that trial and seeded with a SAM2~\citep{sam2} first-frame mask. To consolidate per-view estimates into a single trajectory in the rig frame, we perform a confidence-weighted multi-view fusion using FoundationPose's per-view score, followed by a sliding-window bundle adjustment that minimizes per-view reprojection error together with a temporal-smoothness term on the 6-DoF pose.

\textbf{Hand pose.} We estimate per-view MANO~\citep{mano} parameters with WiLoR~\citep{wilor}, which produces hand pose at unknown scale from a single RGB image. We recover metric MANO predictions by solving for a per-trial global scale and rigid alignment of the WiLoR-predicted 3D hand keypoints to the multi-view RGB-D point cloud and the multi-view-triangulated 2D hand keypoints. We then apply the same multi-view fusion and temporal smoothing as for tool pose to obtain a single MANO trajectory per hand in the rig frame.

\textbf{Joint hand--tool refinement.} Following HO-Cap~\citep{hocap}, we perform a joint hand--tool pose optimization that minimizes a weighted sum of (i) multi-view 2D hand-keypoint reprojection error, (ii) hand-keypoint-to-RGB-D-depth distance, (iii) tool 6-DoF reprojection error against the multi-view masks, and (iv) a signed-distance-field-based contact prior between the MANO mesh and the tool mesh that penalizes interpenetration while encouraging contact during active manipulation phases. Full pipeline details, hyperparameters, and per-component ablations are in the appendix.





\section{Creative Physical Tool-Use Benchmarks}
We establish a suite of creative tool-use benchmarks spanning physical reconstruction, reasoning, and robot learning. We evaluate sparse-view static and dynamic reconstruction with Gaussian Splatting (Section \ref{sec:app-3dgs-main}), benchmark vision-language models on physical tool-use reasoning (Section \ref{sec:vlm_benchmark}), and study robot policy learning from human tool-use demonstrations (Section \ref{sec:robot_tooluse}).

\subsection{Novel-View Gaussian Splatting for Sparse-View Tooluse: Static and Dynamic}
\label{sec:app-3dgs-main}

\paragraph{Experiment setup.} We benchmark sparse-view novel-view synthesis on 8-camera calibrated capture in two regimes: (i) static 3DGS using InstantSplat~\citep{fan2024instantsplat} (IS) as a backbone (ii) dynamic 4DGS training InstantSplat on $T$-frame sequences applied per-timestep, denoted IS-4D. Each scene is reconstructed twice (once with 8 cameras and once with cam$_6$ dropped for ground truth comparison). We add a second stage single-step diffusion to the pipeline utilizing GSFix3D~\citep{yin2025gsfixer} and Difix3D~\citep{wei2025difix3d} to enhance reconstruction and decrease noise.

\paragraph{Method evaluation.}
We also evaluate two-stage recipes:
IS + Difix3D for 3DGS and IS-4D+Difix3D+GSFix3D for 4DGS. \emph{Stage~1 (reconstruction):} a MASt3R~\citep{leroy2024mast3r} cross-view dense-point prior seeds the joint pose-and-Gaussian optimizer of InstantSplat, trained for 1k iterations under a clean photometric loss with no depth/distortion/normal regularizers. For 4DGS, Stage~1 is run independently per timestep. \emph{Stage~2 (rendered-view repair):} a single-step Difix3D~\citep{wei2025difix3d} pass, with GSFix3D~\citep{yin2025gsfixer} chained after Difix3D in the 4DGS variant (More benchmarks in Appendix \ref{sec:app-3dgs}).

\paragraph{Reconstruction results.}
~\autoref{tab:3dgs-4dgs-combined} summarizes both regimes. PSNR (in dB; higher = closer pixel-by-pixel to the held-out cam$_6$ ground truth) measures photometric fidelity, while MUSIQ (on a 0--100 scale; higher = better learned-perceptual appeal) judges visual quality without any reference. \emph{Static:} IS + Difix3D reaches \textbf{20.45}\,PSNR / 55.11\,MUSIQ, beating InstantSplat alone by $+0.5$\,dB and SplatFormer~\citep{chen2025splatformer} by ${\sim}7$\,dB. \emph{Dynamic:} IS-4D + Difix3D + GSFix3D reaches \textbf{69.92}\,MUSIQ ($+18.92$ over IS-4D) while lowering PSNR by 3.6\,dB -- the photometric-vs-perceptual trade-off~\citep{wei2025difix3d} is now measurable on held-out 4D. Both findings are made quantifiable only by our dataset's \emph{calibrated 8-camera capture}, which uniquely supports a held-out 
camera view test on tool-use scenes that competing tabletop datasets cannot reproduce. Without it, neither the static held-out gap nor the 4D photometric-vs-perceptual dissociation can be measured.

\begin{figure}[t]
\centering
\includegraphics[width=\linewidth]{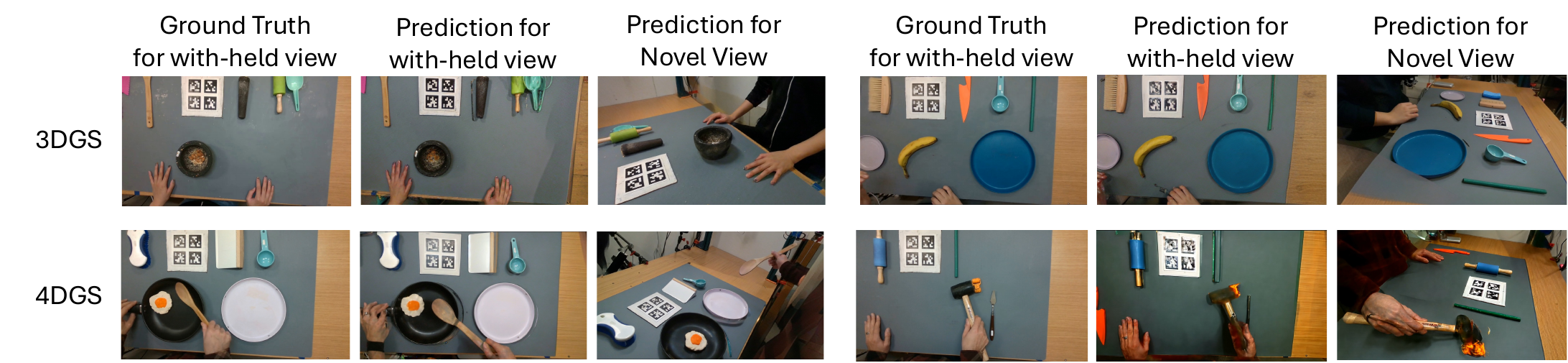}
\caption{3DGS and 4DGS prediction for camera view withheld during training and prediction of novel views using InstantSplat + Difix3D. Additional visualizations are available on our website.
}
\vspace{-0.1in}
\label{fig:reconstruction-prediction}
\end{figure}

\begin{table}[t]
  \centering
  \caption{Novel-view synthesis results for data captured by 8 sparse cameras, for both
  static (3DGS) and dynamic (4DGS) settings. PSNR is computed against a held-out camera
  view (cam$_6$); MUSIQ scores are computed after training with all 8 cameras (no ground
  truth required). Methods: SplatFormer~\citep{chen2025splatformer}, InstantSplat
  (IS)~\citep{fan2024instantsplat}, Difix3D~\citep{wei2025difix3d},
  GSFix3D~\citep{yin2025gsfixer}; MUSIQ~\citep{ke2021musiq}.}
  \label{tab:3dgs-4dgs-combined}
  \small
  \begin{tabular}{@{}lccccc@{}}
    \toprule
    & \multicolumn{3}{c}{\textit{Static (3DGS)}} & \multicolumn{2}{c@{}}{\textit{Dynamic (4DGS)}} \\
    \cmidrule(lr){2-4} \cmidrule(l){5-6}
    & SplatFormer & IS & \textbf{IS+Difix3D} & IS-4D & \textbf{IS-4D+Difix3D+GSFix3D} \\
    \midrule
    PSNR $\uparrow$  & 13.09 & 19.97 & \textbf{20.45} & \textbf{19.60} & 15.97 \\
    MUSIQ $\uparrow$ & 33.11 & 50.66 & \textbf{55.11} & 51.00 & \textbf{69.92} \\
    \bottomrule
  \end{tabular}
  \vspace{-0.24in}
\end{table}

\subsection{Benchmarking VLMs' Physical Tool-use Capabilities}
\label{sec:vlm_benchmark}

\textbf{Experiment setup.} We evaluate three levels of reasoning capabilities. For \textbf{Tool Selection}, models are asked to identify all tools that could plausibly complete the task, and a prediction is counted as correct if any predicted tool matches the demonstrated ground-truth tool. For \textbf{Spatial Grounding}, models localize the selected tool using bounding boxes, evaluated using IoU only on semantically matched tool predictions. 
For \textbf{Behavioral Planning}, models are provided with a natural-language instruction, e.g., ``stir the coffee.'' and must generate a high-level manipulation trajectory based on the first frame from a top-view camera. 
 The evaluation subset contains 180 manually verified trajectories spanning 32 task variants. 
We evaluate a tool-center projection ground-truth trajectory and a correspondence-aware transported-point trajectory, using trajectory distance metrics including Mean L2 and Chamfer-L2. Details are in Appendix~\ref{subsection:benchmarking-details}.

\paragraph{Tool selection and spatial grounding results.} As shown in \autoref{tab:vlm_tool_benchmark}, modern VLMs demonstrate strong capability in selecting plausible tools. Claude-Sonnet-4.6, Claude-Opus-4.7 and GPT-6-Astra achieve the highest tool-selection performance of 80\%, while GPT-6-Astra substantially outperforms all other models on spatial grounding with a bounding box IoU of 0.891. Notably, Gemini and Qwen models frequently correctly identify plausible tools but struggle to obtain correct bounding boxes, suggesting that these VLMs possess substantially stronger semantic reasoning than fine-grained spatial understanding.

\begin{table}[h]
\vspace{-1em} 
\centering
\footnotesize 
\setlength{\tabcolsep}{4pt} 
\renewcommand{\arraystretch}{0.85} 
\caption{Benchmark results on tool selection accuracy and bounding box IoU.
Models: GPT-5-mini ~\citep{openai_gpt5_2025}, GPT-6-Astra~\citep{openai_gpt6_astra_2026},
Claude-Sonnet-4.6~\citep{anthropic_sonnet46_2026},Claude-Opus-4.7~\citep{anthropic_sonnet46_2026}
Gemini-3.1-Pro~\citep{google_gemini3pro_2026},
Qwen-3.6-27B~\citep{qwen36_27b}}
\vspace{-0.8em} 
\label{tab:vlm_tool_benchmark}
\begin{tabular}{lcc}
\toprule
\textbf{Model} & \textbf{Tool Acc.} $\uparrow$ & \textbf{BBox IoU} $\uparrow$ \\ 
\midrule
GPT-5-mini        & .662          & .300 \\
GPT-6-Astra          & \textbf{.800}          & \textbf{.891} \\
Claude-Sonnet-4.6 & \textbf{.800} & .641 \\
Claude-Opus-4.7   & \textbf{.800} & .564 \\
Gemini-3-Flash    & .756          & .050 \\
Gemini-3.1-Pro    & .781          & .010 \\
Qwen-3.5-9B       & .557          & .003 \\
Qwen-3.6-27B      & .638          & .004 \\
\bottomrule
\end{tabular}
\vspace{-1.5em} 
\end{table}


\paragraph{Behavioral planning results.}
We additionally compare several general-purpose reasoning/VLMs with Gemini Robotics ER on zero-shot behavioral planning. As shown in \autoref{tab:behavioral-results}, GPT-6-Astra achieves the highest evaluation-success rate (89.4\%). Among successfully evaluated trajectories Gemini Robotics ER 1.6 \citep{gemini_2023} achieves the lowest trajectory errors under both trajectory constructions. These results highlight a tradeoff between output reliability and model spatial awareness. 


\begin{table}[h]
\centering
\caption{Behavioral-planning results across all 160 trajectories for GPT-6-Astra ~\citep{openai_gpt6_astra_2026}, Gemma 4 31B~\citep{google_gemma4_2026}, Gemini Robotics ER~\citep{gemini_2023}, Claude Fable 5~\citep{anthropic_fable5_2026}, and Qwen 3.6 27B~\citep{qwen36_27b_2026}. All distance metrics are reported in pixels and computed over successfully evaluated outputs.}
\label{tab:behavioral-results}
\vspace{-0.8em}
\footnotesize
\setlength{\tabcolsep}{3.5pt}
\renewcommand{\arraystretch}{1.0}
\begin{tabular}{lccccc}
\toprule
\textbf{Model} &
\makecell{\textbf{Eval.}\\\textbf{success} $\uparrow$} &
\multicolumn{2}{c}{\textbf{Tool-center}} &
\multicolumn{2}{c}{\textbf{Corr.-aware}} \\
\cmidrule(lr){3-4}
\cmidrule(lr){5-6}
&
&
\textbf{Mean L2} $\downarrow$ &
\textbf{Chamfer-L2} $\downarrow$ &
\textbf{Mean L2} $\downarrow$ &
\textbf{Chamfer-L2} $\downarrow$ \\
\midrule
GPT-6 Astra.
    & \textbf{89.4\%}
    & 239.37
    & 120.38
    & 227.85
    & 97.11 \\
Gemma 4 31B
    & 86.9\%
    & 259.96
    & 104.17
    & 269.73
    & 107.31 \\
Gemini Robotics ER
    & 77.5\%
    & \textbf{230.37}
    & \textbf{91.91}
    & \textbf{223.15}
    & \textbf{74.44} \\
Claude Fable 5
    & 71.3\%
    & 263.33
    & 146.37
    & 234.89
    & 102.61 \\
Qwen 3.6 27B
    & 37.5\%
    & 244.88
    & 128.44
    & 228.61
    & 103.71 \\
\bottomrule

\end{tabular}
\vspace{-1.5em}
\end{table}


\paragraph{Qualitative results.}

\begin{figure}[h]
\centering
\includegraphics[width=1\textwidth]{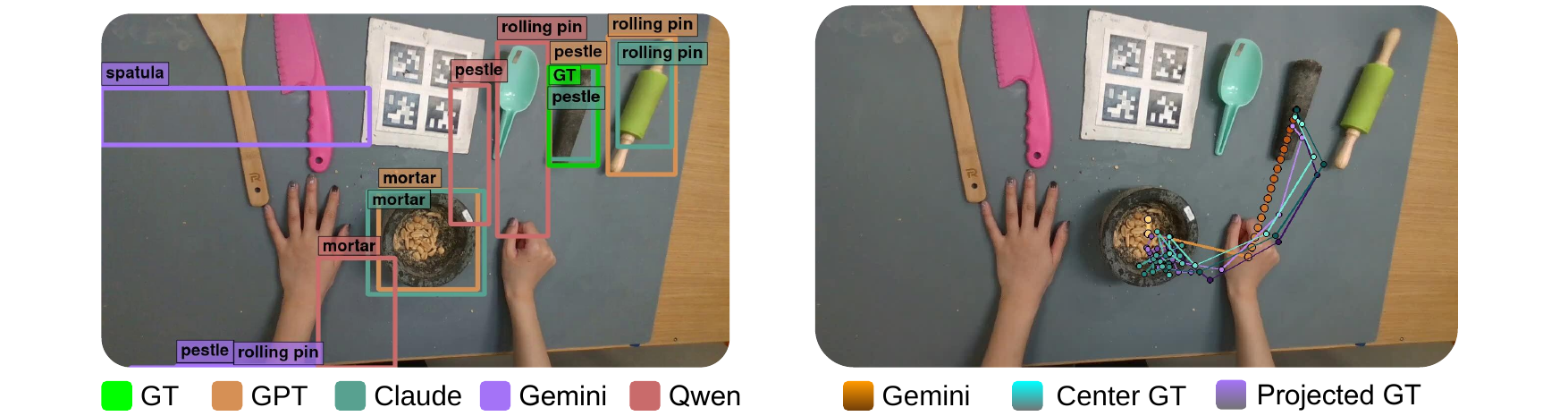}
\caption{
\emph{Left:} Predicted tool names and their corresponding bounding boxes generated by GPT 6 Astra, Claude Sonnet 4.6, Gemini 3.1 Pro, and Qwen3.6-27B. 
\emph{Right:} Ground-truth task trajectories together with trajectories predicted by Gemini Robotics ER 1.6.
}
\vspace{-0.2in}
\label{fig:vlm-results}
\end{figure}

Qualitatively, many VLMs demonstrate creative physical reasoning beyond reproducing the canonical ground-truth tool. As shown in \autoref{fig:vlm-results}, several models identify the rolling pin as a plausible alternative to the mortar and pestle for crushing nuts, while GPT-6-Astra produces near ground-truth spatial grounding. Additional examples are provided in Appendix~\ref{subsection:spatial-appendix}.
Predicted trajectories generally capture the intended manipulation behavior for short-horizon tasks such as pushing and cutting, but are less consistent for longer-horizon actions requiring repeated interactions such as sweeping or shaping. In \autoref{fig:vlm-results}, Gemini Robotics ER correctly approaches the mortar and pestle but fails to capture the repeated grinding motion needed to crush the nuts. Additional qualitative examples are provided in Appendix~\ref{subsection:gemini-appendix}. These results highlight the capabilities and limitations of modern VLMs, demonstrating that \dataname ~ provides a challenging benchmark for evaluating semantic understanding, spatial grounding, and task planning in real-world settings.

\subsection{Robot Physical Tool-Use from Human Demonstrations}
\label{sec:robot_tooluse}
We benchmark representative paradigms for transferring human tool-use demonstrations to dexterous robot policies, ranging from direct motion transfer to interaction-aware reinforcement learning. \emph{Kinematic Retarget} directly executes retargeted human motion, while \emph{BC (Retarget)} learns from the retargeted trajectories. \emph{ManipTrans}~\cite{li2025maniptransefficientdexterousbimanual} uses residual reinforcement learning to track demonstrated interactions, and \emph{Dex-X (RL)}~\cite{chen2026dexxlearningvisualtactiledexterous} jointly adapts hand and arm motion through physical interaction.

\paragraph{Results.}
\autoref{tab:policy_results} summarizes the task success rates. Direct motion transfer performs poorly under physical interaction: Kinematic Retarget achieves 14.0\% average success, while BC reaches 4.6\%. In contrast, reinforcement-learning-based methods perform substantially better, with ManipTrans achieving 40.5\% and Dex-X RL 55.1\%. The gap is especially pronounced on contact-intensive tasks: pestle crushing improves from 16.8\% to 46.7\%, and squeegee collecting from 12.6\% to 57.5\%. Performance also varies considerably across tools for the same action, indicating that tool geometry and contact structure strongly affect transfer difficulty.

\begin{table}[t]
\centering
\small
\caption{Task success rates (\%) on the benchmark. Higher is better.}
\label{tab:policy_results}
\begin{tabular}{@{}llrrrr@{}}
\toprule
Action & Tool &
\shortstack{Kinematic\\Retarget} &
\shortstack{BC\\(Retarget)} &
\shortstack{ManipTrans\\(RL)} &
\shortstack{Dex-X\\(RL)} \\
\midrule

Stir coffee
& spatula
& 7.0
& 1.6
& 52.2 $\pm$ 2.2
& \textbf{72.7 $\pm$ 0.8} \\

Scoop nuts
& ladle
& 22.2
& 9.1
& 62.9 $\pm$ 0.4
& \textbf{66.8 $\pm$ 1.3} \\

Crush banana
& mallet
& 10.2
& 2.6
& 53.0 $\pm$ 1.9
& \textbf{55.8 $\pm$ 0.2} \\

Scoop nuts
& scooper
& 31.0
& 9.8
& \textbf{45.3 $\pm$ 2.0}
& 31.3 $\pm$ 0.6 \\

Crush nuts
& pestle
& 7.0
& 1.9
& 16.8 $\pm$ 0.2
& \textbf{46.7 $\pm$ 2.4} \\

Collect sand
& squeegee
& 6.9
& 2.6
& 12.6 $\pm$ 0.7
& \textbf{57.5 $\pm$ 1.9} \\

\midrule
\textbf{Average}
& --
& \textbf{14.0}
& \textbf{4.6}
& \textbf{40.5}
& \textbf{55.1} \\

\bottomrule
\end{tabular}
\vspace{-0.25in}
\end{table}

\vspace{-0.1in}
\paragraph{Findings.}
The benchmark highlights three main challenges in human-to-robot tool-use transfer: contact-rich interactions remain difficult, tool geometry can substantially affect performance even within the same action category, and embodiment mismatch limits direct transfer across human and robot morphologies. These results motivate further study of cross-tool generalization, embodiment-aware policy learning, and robust transfer from human demonstrations to physical robot interaction.

 
\section{Conclusion}

We introduced \dataname, a large-scale ego-exo human demonstration dataset and benchmark for creative physical tool use. 
\dataname~explicitly captures the full \textbf{hand--tool--target} interaction in real-world manipulation scenarios, including challenging deformable target objects. 
With multi-view observations, the dataset provides rich annotations for studying embodied physical tool-use.
Beyond the dataset itself, we establish a unified benchmark spanning physical tool-use reasoning, 3D and 4D reconstruction, and robot policy learning. 
Meanwhile, the dataset may raise concerns related to the misuse of human behavior data or unsafe downstream robotic deployment. To mitigate these, all data are collected with consent and released for research purposes under a non-commercial license. 
We hope \dataname~will serve as a foundation for future research on developing robots capable of adaptive tool use in open-world environments. Limitations are discussed in Appendix \ref{sec:limitations}








\bibliographystyle{iclr2027_conference}

\subsection*{AI use statement}


Large language models were used throughout this work under author direction: for literature collection, experimental statistics, code development, and manuscript drafting, with additional assistance in
experimental design. All core ideas, the experimental program, and the conclusions originate from
the authors. The authors verified the paper’s numerical claims against the underlying configurations,
logs, evaluation records, and analysis artifacts, reviewed all AI-assisted text, code, and figures, and
take full responsibility for the paper’s content.


\clearpage
\newpage
\appendix

\section{Appendix}



\subsection{Novel-View Gaussian Splatting for Sparse-View Tooluse: Static and Dynamic}
\label{sec:app-3dgs}

\paragraph{Experiment Setup.} We benchmark sparse-view novel-view synthesis in two regimes that share a single calibrated capture: \emph{static} 3DGS on a per-scene image, and \emph{dynamic} 4DGS on a per-scene $T$ sequence. In both cases we use the same 8 calibrated RealSense cameras and the same held-out cam$_6$ Protocol~B -- each scene is reconstructed twice: an \emph{8-cam} run that uses all eight cameras for training and a \emph{cam$_6$-dropped} run that uses cam$_0,\dots,5,7$ for training and reserves cam$_6$ for held-out evaluation. From the cam$_6$-dropped run we score the predicted cam$_6$ image (or render across time at cam$_6$'s pose, in 4D) against the captured ground truth with full-reference PSNR, SSIM~\citep{wang2004ssim}, and LPIPS~\citep{zhang2018lpips}. From the 8-cam run we score the rendered novel-view tour without a reference using the appeal-composite metrics MUSIQ~\citep{ke2021musiq}, MANIQA~\citep{yang2022maniqa}, TOPIQ~\citep{chen2024topiq}, BRISQUE~\citep{mittal2012brisque}, NIQE~\citep{mittal2013niqe}, and PIQE~\citep{venkatanath2015piqe}. Baselines use their native initialization: SparseGS reads the COLMAP~\citep{schoenberger2016mvs,schoenberger2016sfm} sparse reconstruction, while Sparfels, SplatFormer (applied to our InstantSplat model), and InstantSplat are seeded from MASt3R~\citep{leroy2024mast3r}. We use InstantSplat~\citep{fan2024instantsplat} as our main backbone; in the dynamic regime we apply InstantSplat \emph{across the $T$ timesteps}, which we refer to as InstantSplat-4D.

\begin{figure}[h]
\centering
\includegraphics[width=0.8\linewidth]{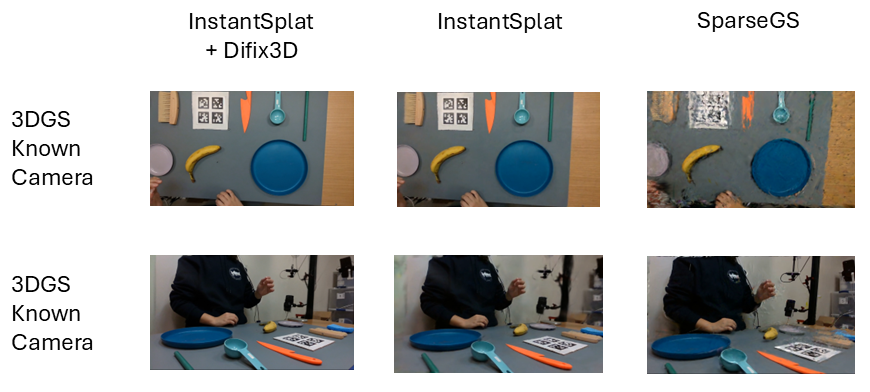}
\caption{Training of 3DGS using 8 cameras and reconstruction of a known camera view
}
\vspace{-0.1in}
\label{fig:qual-1}
\end{figure}

\begin{figure}[h]
\centering
\includegraphics[width=0.8\linewidth]{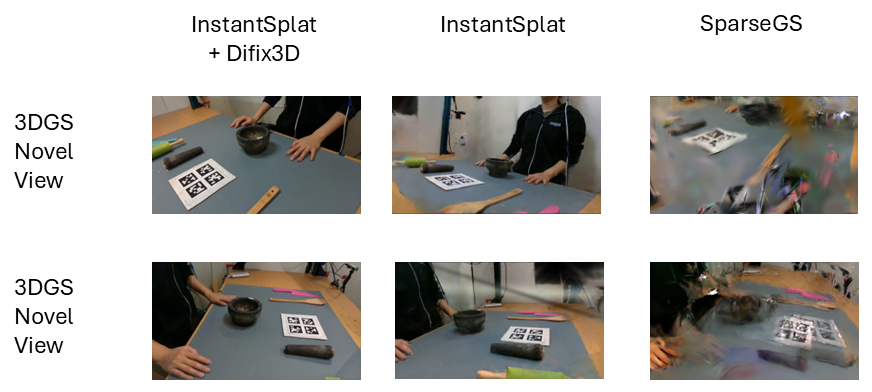}
\caption{Training of 3DGS using 8 cameras and reconstruction of a novel camera view
}
\vspace{-0.1in}
\label{fig:qual-2}
\end{figure}

\paragraph{Method.}
We propose a two-stage sparse-view recipe specifically engineered for the failure modes of unposed tabletop tool-use reconstruction, taking the seven (or eight) training views of our calibrated capture as input -- and, in the dynamic case, applied across $T$ timesteps. \emph{Stage~1 -- reconstruction.} We obtain a geometry-aware initialization from a MASt3R~\citep{leroy2024mast3r} cross-view dense point prior and train the joint pose-and-Gaussian optimizer of InstantSplat~\citep{fan2024instantsplat} for 1k iterations under a clean photometric loss; we deliberately omit every reconstruction-time regularizer (depth, distortion, normals) that we found to leave held-out PSNR flat or harm SSIM/LPIPS in our setting. For 4DGS we instantiate an independent Gaussian field $\mathbb{G}_t$ at each timestep $t$ using the same Stage-1 design, with the MASt3R prior recomputed per timestep. \emph{Stage~2 -- rendered-view repair.} Rather than expending optimizer budget on 3D losses that do not address the dominant sparse-cam failure mode, we move the heavy lifting to the rendered-view level where ghosting and missing high-frequency detail actually live. For 3DGS we apply a single-step diffusion utilizing Difix3D~\citep{wei2025difix3d} repair tuned to the rendering distribution of tool-use scenes; for 4DGS we additionally chain GSFix3D~\citep{yin2025gsfixer} after Difix3D.

\begin{table*}[t]
  \centering
  \caption{3DGS novel-view synthesis with 8 sparse cameras, benchmarked on three tabletop
  tool-use scenes (\textit{pestle\_crush\_nuts}, \textit{ladle\_scoop\_nuts},
  \textit{knife\_slice\_banana}). Each entry is the mean over the three scenes.
  Methods: SplatFormer~\citep{chen2025splatformer}, Sparfels~\citep{jena2025sparfels},
  SparseGS~\citep{xiong2024sparsegs}, InstantSplat~\citep{fan2024instantsplat}.}
  \label{tab:3dgs-main}
  \footnotesize
  \setlength{\tabcolsep}{3pt}
  \begin{tabular}{@{}lccccccccc@{}}
    \toprule
    & \multicolumn{3}{c}{\textbf{Full-reference}} & \multicolumn{6}{c@{}}{\textbf{No-reference appeal composite}} \\
    \cmidrule(lr){2-4} \cmidrule(l){5-10}
    Method & PSNR$\uparrow$ & SSIM$\uparrow$ & LPIPS$\downarrow$
           & MUSIQ$\uparrow$ & MANIQA$\uparrow$ & TOPIQ$\uparrow$
           & BRISQUE$\downarrow$ & NIQE$\downarrow$ & PIQE$\downarrow$ \\
    \midrule
    SplatFormer  & 13.09 & 0.492 & 0.694 & 33.11 & 0.255 & 0.370 & 58.38 & 5.72 & 44.95 \\
    Sparfels     & 13.72 & 0.534 & 0.642 & \textbf{61.16} & \textbf{0.331} & 0.436 & 59.39 & 6.27 & 67.68 \\
    SparseGS     & 14.64 & 0.630 & 0.645 & 37.61 & 0.189 & 0.254 & 45.85 & 5.57 & 70.99 \\
    InstantSplat & 19.97 & 0.769 & 0.428 & 50.66 & 0.244 & 0.403 & 38.34 & \textbf{4.56} & 49.93 \\
    \midrule
    \textbf{IS + Difix3D} & \textbf{20.45} & \textbf{0.779} & \textbf{0.364} & 55.11 & 0.324 & \textbf{0.456} & \textbf{21.15} & 4.81 & \textbf{25.13} \\
    \bottomrule
  \end{tabular}
\end{table*}

\paragraph{Findings (Static, 3DGS).}
The defining property of our dataset is a calibrated 8-camera capture per scene, which uniquely supports a held-out-cam$_6$ full-reference evaluation that competing tabletop tool-use datasets cannot reproduce -- and \autoref{tab:3dgs-main} shows what this enables. Across three tool-use scenes our recipe reaches \textbf{20.45}\,PSNR / \textbf{0.779}\,SSIM / \textbf{0.364}\,LPIPS on cam$_6$, beating InstantSplat by $+0.5$\,dB and Sparfels / SplatFormer / SparseGS by $5\text{--}8$\,dB, while also winning BRISQUE, TOPIQ, and PIQE on the same predicted image. Sparfels takes the top MUSIQ (61.16) because its surfel rasterization produces smooth, locally pleasing texture, but its low full-reference numbers show that clean appearance need not imply faithful geometry. The headline result is that high-quality sparse-view 3DGS for tool-use scenes emerges only when both sides are in place: our calibrated capture (\textit{dataset}) and our two-stage pipeline (\textit{method}). Without the dataset, neither the held-out gap nor its closure would be measurable; without the method, the gap remains.

\begin{figure}[h]
\centering
\includegraphics[width=0.8\linewidth]{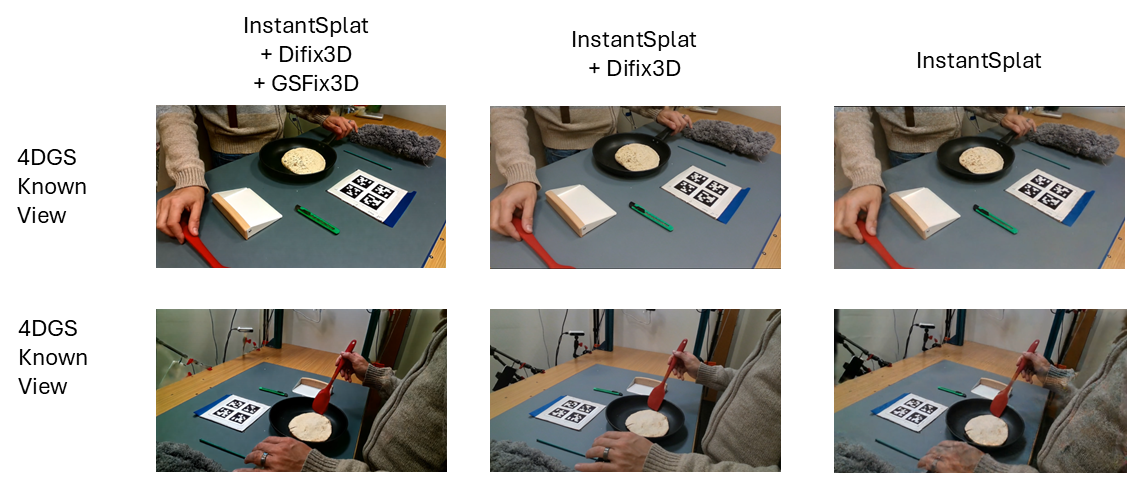}
\caption{Training of 4DGS using 8 cameras and reconstruction of a known camera view
}
\vspace{-0.1in}
\label{fig:qual-3}
\end{figure}

\begin{figure}[h]
\centering
\includegraphics[width=0.8\linewidth]{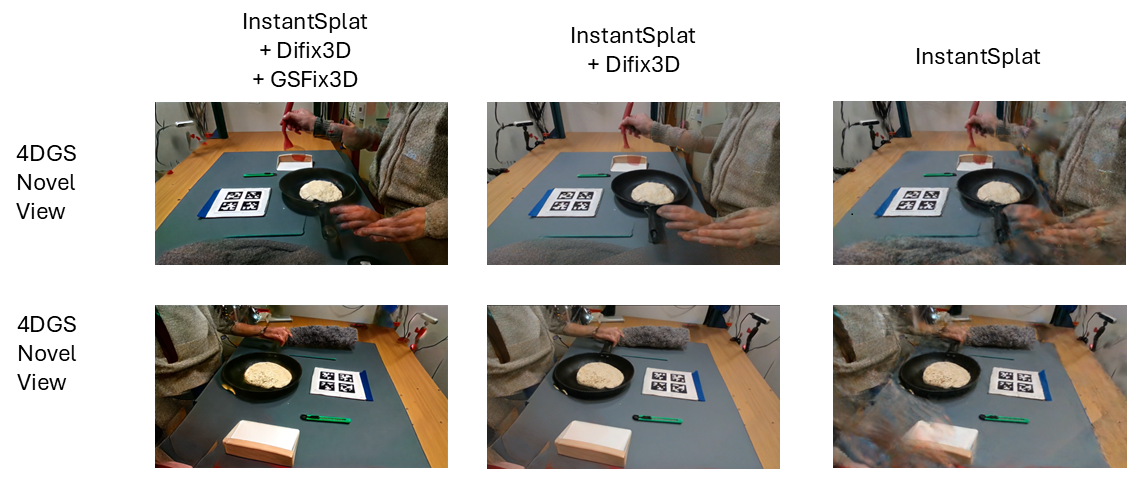}
\caption{Training of 4DGS using 8 cameras and reconstruction of a novel camera view
}
\vspace{-0.1in}
\label{fig:qual-4}
\end{figure}

\paragraph{Findings (Dynamic, 4DGS).}
The same dataset property extends naturally to the dynamic regime, supporting a held-out-cam$_6$ full-reference evaluation \emph{at every timestep} that competing tabletop dynamic-tooluse datasets cannot reproduce. \autoref{tab:4dgs-main} shows across three tool-use scenes our final pipeline reaches \textbf{69.92}\,MUSIQ / \textbf{0.39}\,MANIQA / \textbf{0.54}\,TOPIQ on the 8-cam novel-view tour, and \textbf{4.35}\,NIQE on the same render -- a $+18.92$\,MUSIQ jump over the raw InstantSplat-4D backbone. The full-reference columns reveal the photometric/perceptual dissociation we observed in the static case, now sharpened and localized to the second repair stage: the single-step Difix3D pass leaves held-out fidelity intact (19.60$\,\to\,$19.63\,PSNR, with the best SSIM and LPIPS of any stage), whereas chaining GSFix3D \emph{lowers} PSNR by $3.6$\,dB and SSIM by $0.08$ relative to the raw backbone (19.60$\,\to\,$15.97\,PSNR, 0.715$\,\to\,$0.631\,SSIM) while \emph{raising} every no-reference appeal metric by 19--36\%. This is the well-known photometric-vs-perceptual trade-off~\citep{wei2025difix3d} now measured directly on a held-out 4D novel-view, Our method utilizing Difix3D and GSFix3D rewire noisy per-pixel detail into smoother, plausible-looking content that does not pixel-match the GT but scores higher on learned-perceptual judges. The drop6 protocol is what makes this trade-off measurable in 4D at all; without our calibrated capture, neither the held-out 4D gap nor the photometric-vs-perceptual dissociation would be quantifiable. The headline result mirrors the static case: high-quality sparse-view dynamic novel-view synthesis emerges only when both sides are in place: the 4D-extended calibrated capture (\textit{dataset}) and the two-stage pipeline (\textit{method}).

\begin{table*}[h]
  \centering
  \caption{4DGS Novel-view synthesis with Sparse 8 cameras benchmarked on three dynamic tabletop tool-use scenes
  (\textit{mallet\_push\_toy}, \textit{spatula\_flip\_naan}, \textit{spoon\_lift\_place\_egg}).
  Full-reference metrics score the held-out cam$_6$ render of the cam$_6$-dropped run against its captured
  ground truth at 10 sampled timesteps per scene; no-reference metrics score  rendered novel-view tour
  of the 8-cam run.
}
  \label{tab:4dgs-main}
  \footnotesize
  \setlength{\tabcolsep}{4pt}
  \resizebox{\textwidth}{!}{%
  \begin{tabular}{l ccc | cccccc}
    \toprule
    & \multicolumn{3}{c}{\textbf{Full-reference (held-out cam$_6$)}} & \multicolumn{6}{c}{\textbf{No-reference appeal composite (novel-view tour)}} \\
    \cmidrule(lr){2-4} \cmidrule(lr){5-10}
    Pipeline stage & PSNR$\uparrow$ & SSIM$\uparrow$ & LPIPS$\downarrow$
           & MUSIQ$\uparrow$ & MANIQA$\uparrow$ & TOPIQ$\uparrow$
           & BRISQUE$\downarrow$ & NIQE$\downarrow$ & PIQE$\downarrow$ \\
    \midrule
    InstantSplat-4D
           & 19.60          & 0.715         & 0.372
           & 51.00          & 0.22          & 0.37          & 39.22          & 4.87          & \textbf{48.05} \\
    IS-4D + Difix3D
           & \textbf{19.63} & \textbf{0.728} & \textbf{0.313}
           & 62.34          & 0.32          & 0.44          & 30.39          & 4.84          & 48.13 \\
    IS-4D + Difix3D + GSFix3D
           & 15.97          & 0.631         & 0.352
           & \textbf{69.92} & \textbf{0.39} & \textbf{0.54} & \textbf{29.55} & \textbf{4.35} & 53.33 \\
    \bottomrule
  \end{tabular}%
  }
\end{table*}

\paragraph{Expanded observed-view evaluation (Dynamic, 4DGS).}
Complementing the three-scene held-out protocol above, we additionally ran the full 4D pipeline on 20 dynamic tool-use scenes with \emph{all 8 cameras} used for training. In this setting the full-reference metrics score each of the 8 observed camera views against its own captured ground-truth frame, and the no-reference metrics again score the rendered novel-view tour (\autoref{tab:4dgs-observed}). The raw InstantSplat-4D backbone reconstructs the observed views at \textbf{36.97}\,PSNR / \textbf{0.975}\,SSIM / \textbf{0.044}\,LPIPS, and the Difix3D$\,\to\,$GSFix3D repair stack lifts every no-reference metric (MUSIQ 48.79$\,\to\,$64.69, MANIQA 0.209$\,\to\,$0.388, TOPIQ 0.354$\,\to\,$0.528, with the best BRISQUE, NIQE, and PIQE) while lowering observed-view fidelity (36.97$\,\to\,$21.11\,PSNR) -- the same photometric-vs-perceptual dissociation as above, now measured over 20 scenes.

\begin{table*}[h]
  \centering
  \caption{4DGS evaluation with all 8 cameras observed, averaged over 20 dynamic tool-use scenes.
  Full-reference metrics compare each of the 8 training views with its captured ground truth at up to
  10 sampled timesteps; no-reference metrics score the rendered novel-view tour.}
  \label{tab:4dgs-observed}
  \footnotesize
  \setlength{\tabcolsep}{4pt}
  \resizebox{\textwidth}{!}{%
  \begin{tabular}{l ccc | cccccc}
    \toprule
    & \multicolumn{3}{c}{\textbf{Full-reference (8 observed views)}} & \multicolumn{6}{c}{\textbf{No-reference appeal composite (novel-view tour)}} \\
    \cmidrule(lr){2-4} \cmidrule(lr){5-10}
    Pipeline stage & PSNR$\uparrow$ & SSIM$\uparrow$ & LPIPS$\downarrow$
           & MUSIQ$\uparrow$ & MANIQA$\uparrow$ & TOPIQ$\uparrow$
           & BRISQUE$\downarrow$ & NIQE$\downarrow$ & PIQE$\downarrow$ \\
    \midrule
    InstantSplat-4D
           & \textbf{36.97} & \textbf{0.975} & \textbf{0.044}
           & 48.79 & 0.209 & 0.354 & 42.13 & 5.10 & 50.61 \\
    \quad + Difix3D
           & 31.87 & 0.931 & 0.095
           & 61.93 & 0.311 & 0.444 & 39.25 & 5.23 & 55.66 \\
    \quad + Difix3D + GSFix3D
           & 21.11 & 0.758 & 0.186
           & \textbf{64.69} & \textbf{0.388} & \textbf{0.528} & \textbf{37.78} & \textbf{4.71} & \textbf{49.88} \\
    \bottomrule
  \end{tabular}%
  }
\end{table*}

\subsection{VLM Benchmarking Details}
\label{subsection:benchmarking-details}

\paragraph{Prompt Used}
We use the prompts in \autoref{tab:tool-selection-prompt} for tool selection and spatial grounding, while we used the prompts \autoref{tab:trajectory-prompt} for behavioral trajectory generation.

\begin{table*}[ht]
\setlength\tabcolsep{0pt}
\centering
\begin{tabular*}{\linewidth}{@{\extracolsep{\fill}} l }
\toprule

\begin{lstlisting}[style=promptstyle]
System Prompt:
You are a robot tool-use reasoning assistant. Select which visible tools could realistically complete the task. Do not default to the most typical tool.

User Prompt:
Given the scene with multiple laid-out tools, identify which of them could be used to complete the task: "<TASK>".

Only select tools that are both visible and plausible for completing the task. Unconventional tools are allowed, but do not include tools that clearly would not work.

# only added for Spatial Grounding Tasks
Return one bounding box for each tool in the same order as the tools list. Use pixel coordinates [x1, y1, x2, y2] around the visible tool, with x2 and y2 as the bottom-right exclusive coordinates.

Return only the canonical base tool name.
Do NOT include adjectives, materials, colors, or sizes.

Return ONLY the requested JSON object using this schema:

{
  "tools": ["<exact_tool_name>", "..."],'
  # only added for Spatial Grounding Tasks
  "bboxes": [[0, 0, 0, 0]]
}
\end{lstlisting}

\\
\bottomrule
\end{tabular*}
\caption{Prompt used for evaluating tool selection and localization capabilities of VLMs.}
\label{tab:tool-selection-prompt}

\end{table*}

\begin{table*}[ht]
\setlength\tabcolsep{0pt}
\centering
\begin{tabular*}{\linewidth}{@{\extracolsep{\fill}} l }
\toprule

\begin{lstlisting}[style=promptstyle]
System Prompt:
You are a robot trajectory planning assistant. Reply with JSON only.

User Prompt:
Plan a 2D image-space trajectory for this manipulation task:
"<TASK>"

Place point 0 on the <TOOL_NAME>. Then return <NUM_POINTS>
additional points that describe a plausible trajectory for
using the tool to complete the task.

Use normalized image coordinates in [y, x] format, where each
coordinate is an integer in the range 0-1000.

Return ONLY a JSON array in the following format:

[
  {"point": [y, x], "label": "0"},
  {"point": [y, x], "label": "1"}
]

Requirements:
- Labels must be strings from "0" through "<NUM_POINTS>".
- Include exactly one point for each label.
- Do not include code fences or explanatory text.
\end{lstlisting}

\\
\bottomrule
\end{tabular*}
\caption{Prompt used for behavioral planning and trajectory prediction.}
\label{tab:trajectory-prompt}

\end{table*}

\paragraph{VLM-source} We use the OpenAI API, Anthropic Claude API, and Google Gemini API to evaluate proprietary models. Open-weight Qwen models are served using the vLLM \cite{kwon2023efficient} inference framework. For the evaluated Claude models, we enable extended thinking with a maximum budget of 4096 thinking tokens.

\subsection{Additional VLM outputs}
\label{subsection:spatial-appendix}
\begin{figure*}[t]
\centering

\begin{subfigure}{0.32\textwidth}
    \includegraphics[width=\linewidth]{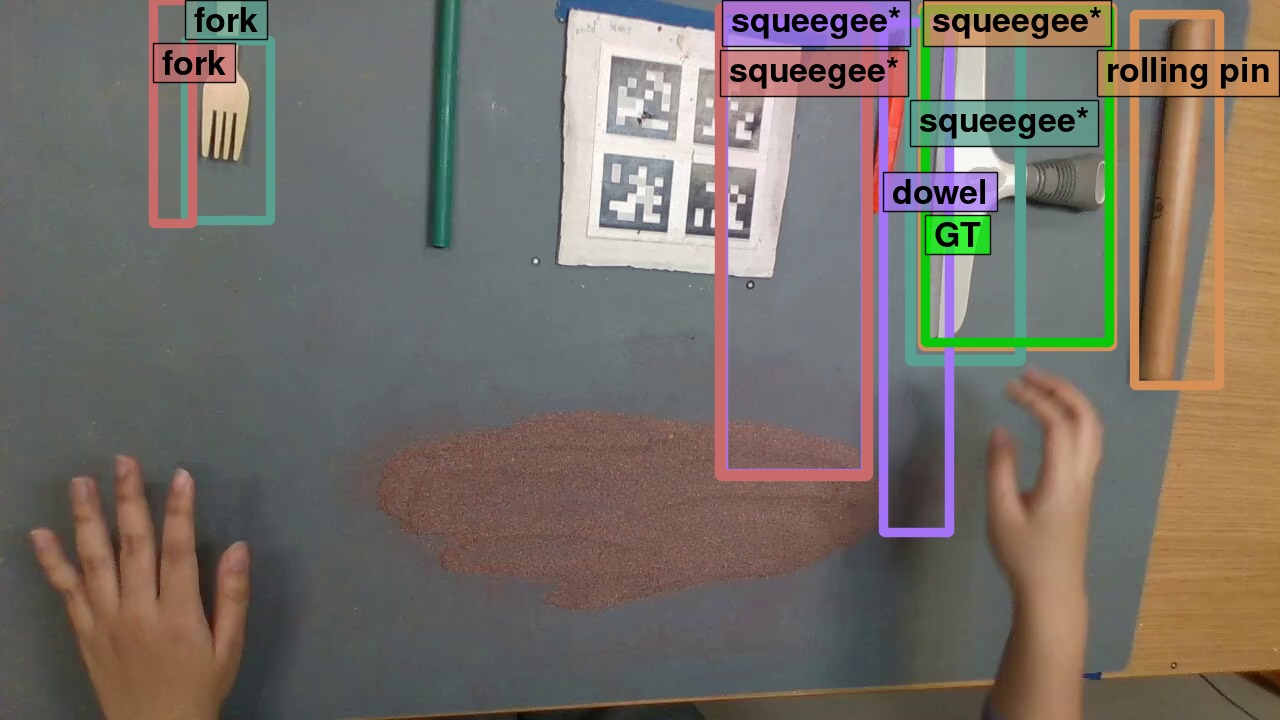}
    \caption{Collect sand.}
\end{subfigure}
\hfill
\begin{subfigure}{0.32\textwidth}
    \includegraphics[width=\linewidth]{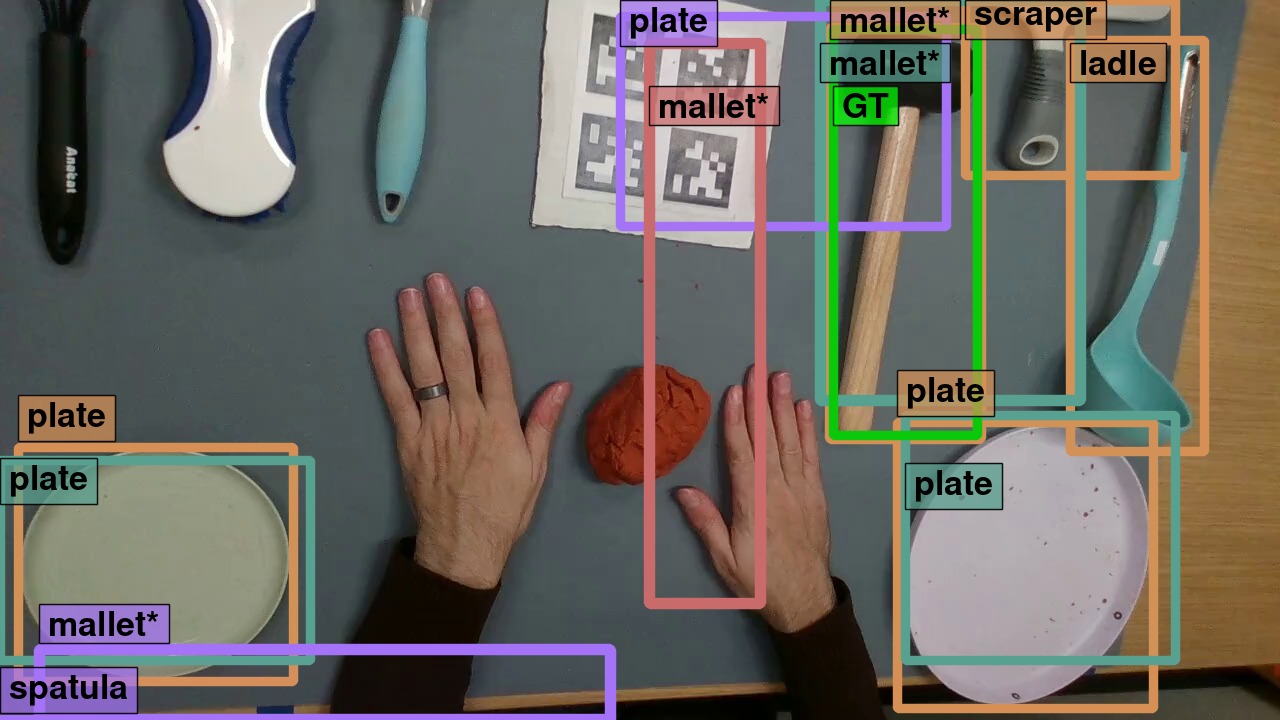}
    \caption{Flatten dough.}
\end{subfigure}
\hfill
\begin{subfigure}{0.32\textwidth}
    \includegraphics[width=\linewidth]{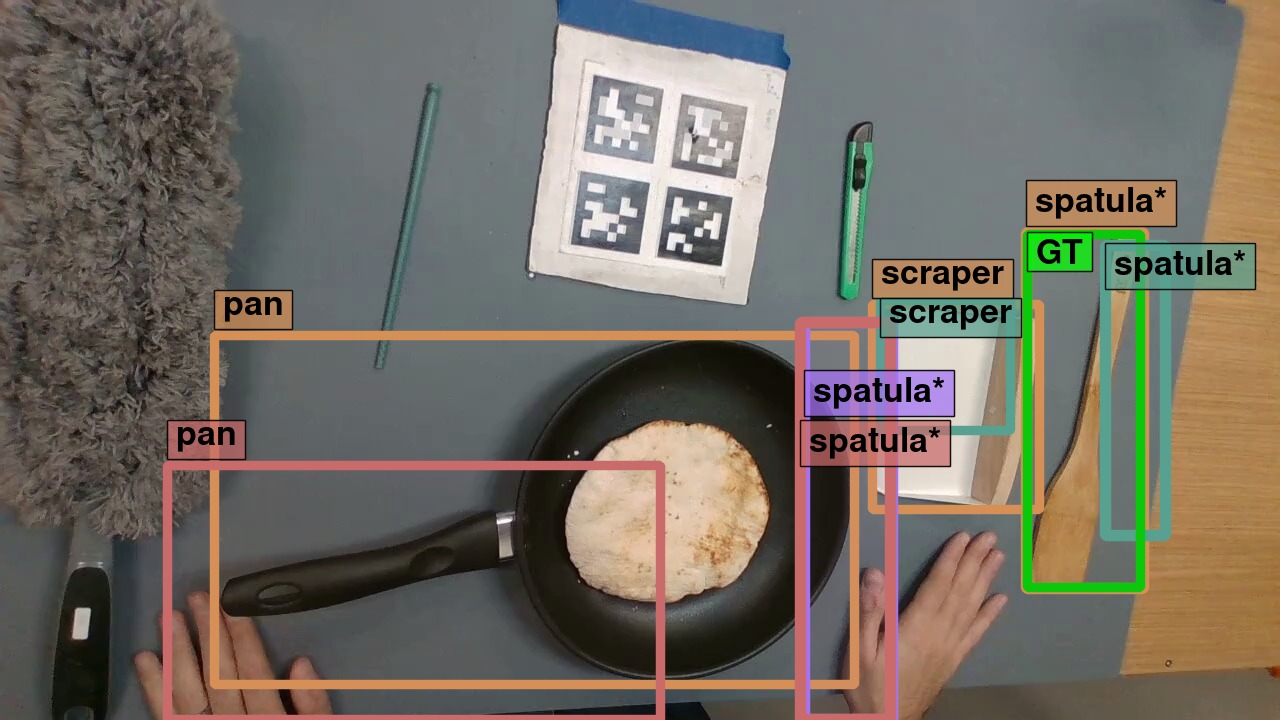}
    \caption{Flip naan.}
\end{subfigure}

\vspace{0.5em}

\begin{subfigure}{0.32\textwidth}
    \includegraphics[width=\linewidth]{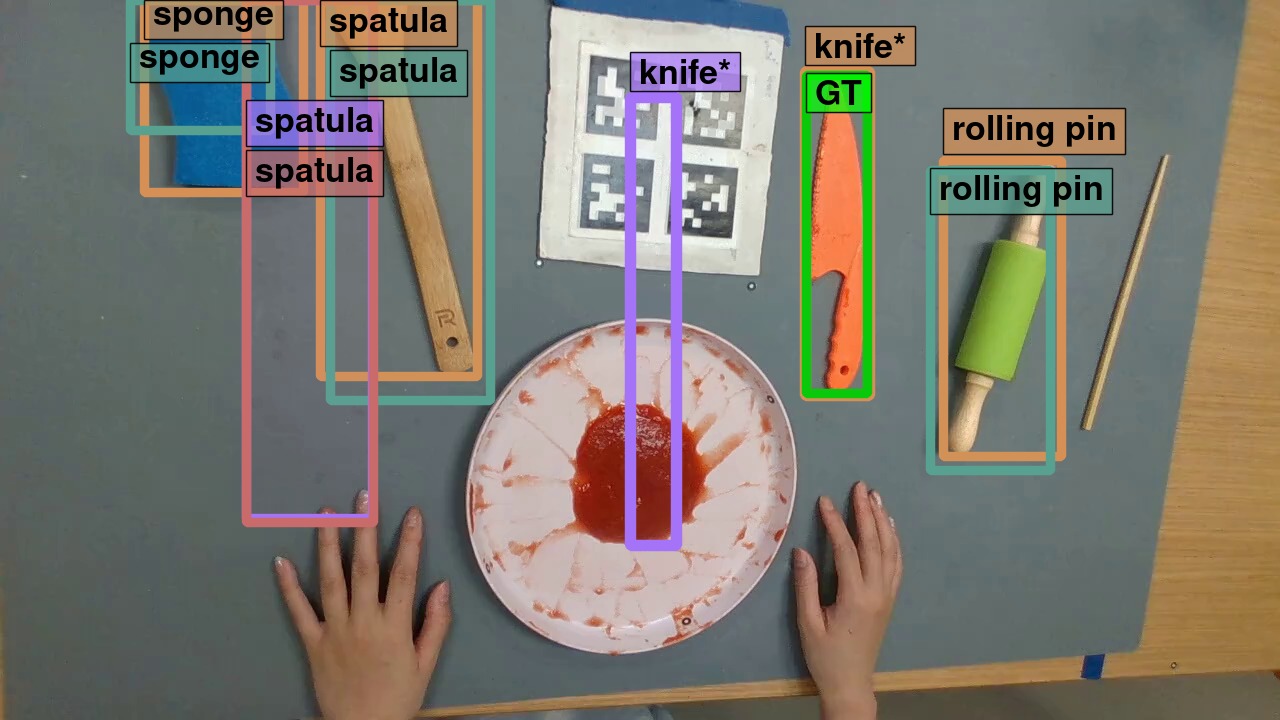}
    \caption{Spread tomato sauce.}
\end{subfigure}
\hfill
\begin{subfigure}{0.32\textwidth}
    \includegraphics[width=\linewidth]{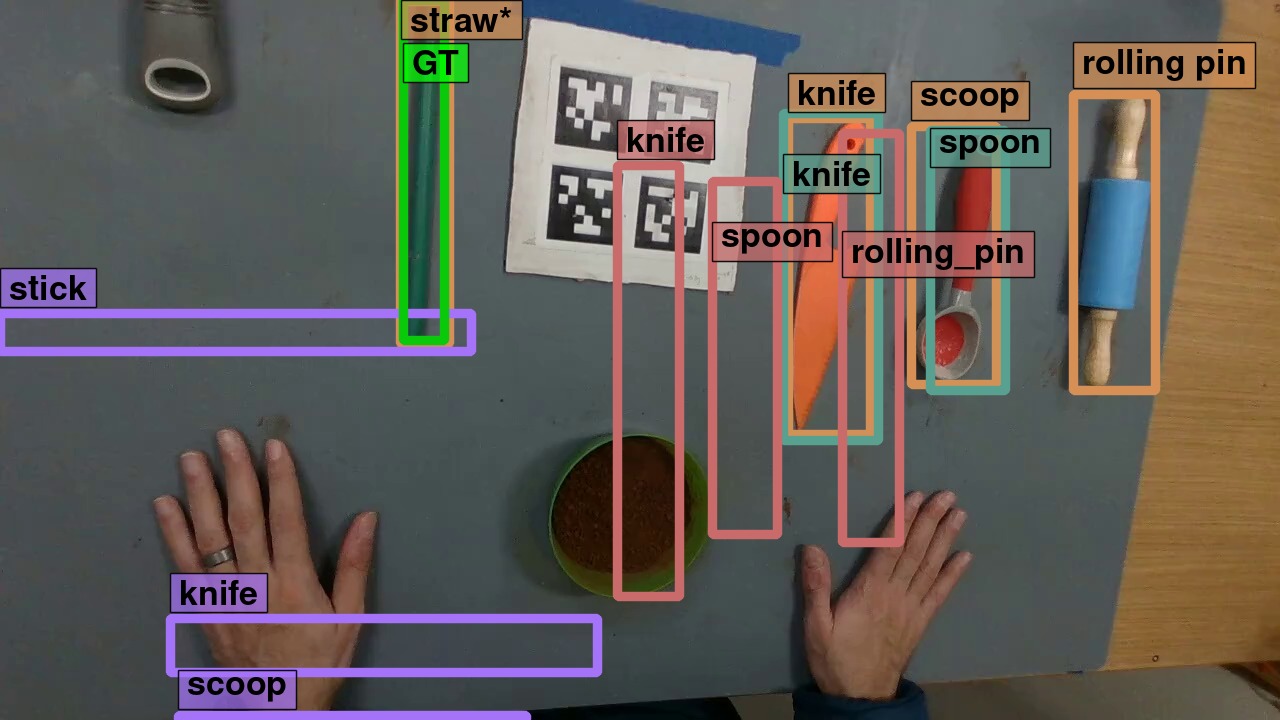}
    \caption{Stir Coffee.}
\end{subfigure}
\hfill
\begin{subfigure}{0.32\textwidth}
    \includegraphics[width=\linewidth]{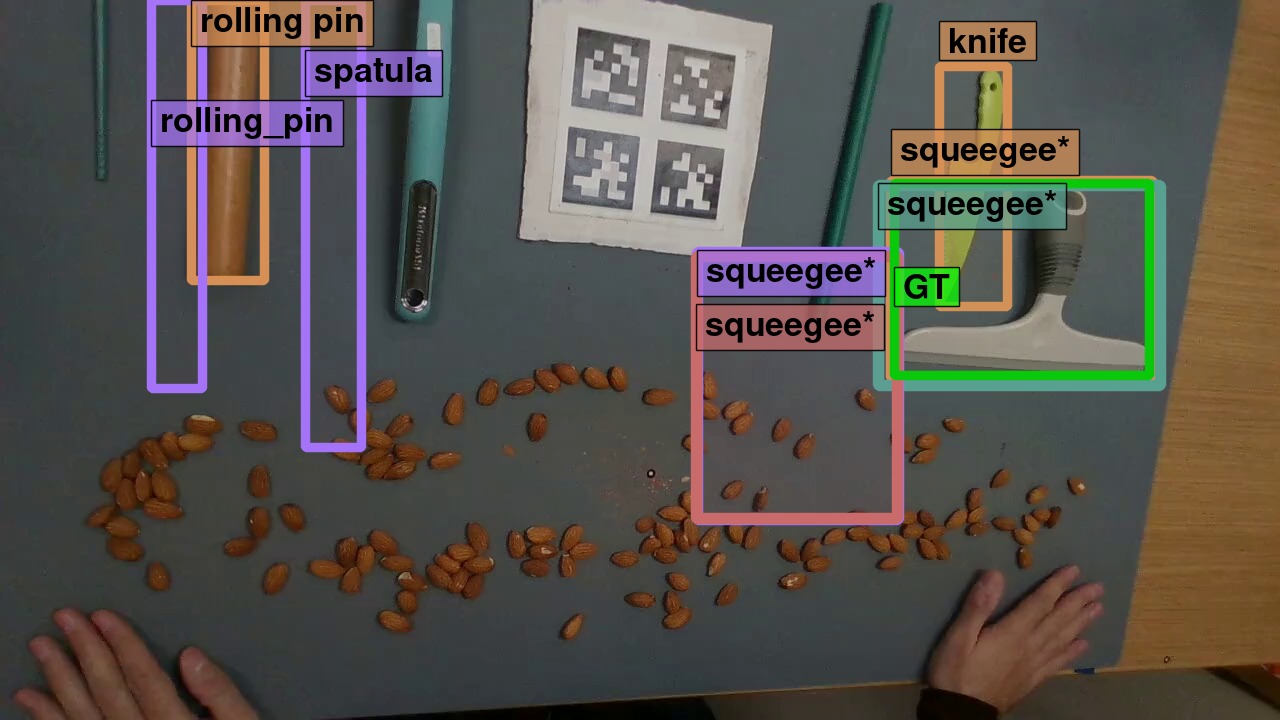}
    \caption{Sweep nuts.}
\end{subfigure}

\caption{
Qualitative trajectory generation results. Bounding boxes are color-coded by model: brown for GPT-6-Astra, teal for Claude Sonnet 4.5, purple for Gemini 3.1 Pro, and pink for Qwen3.6-27B.
}

\label{robotics-vlm-res}
\end{figure*}
Qualitatively, the different VLM families exhibit distinct behaviors. Qwen generally predicts fewer bounding boxes and often defaults to the most common tool associated with a task. While it demonstrates some understanding of spatial relationships, its localization accuracy is weaker; for example, in (a) it correctly reasons that the fork is left of the skewer, but the predicted bounding box is slightly offset. 

GPT models consistently produce the highest-quality spatial grounding results, with bounding boxes frequently overlapping closely with ground-truth annotations. Though occasionaly it doesn't recognize or find smaller objects, such as in (e). Gemini models are notably creative and often generate plausible alternative tool choices, such as suggesting a scoop for stirring coffee. However, Gemini occasionally exhibits weaker spatial understanding, sometimes predicting long horizontal boxes for vertically oriented objects. Claude models are generally the most consistent, frequently producing the correct outputs with relatively few hallucinations, although occasional semantic mistakes still occur, such as identifying a rolling pin as tongs.

\subsection{Gemini Robotics Evaluation and Result Details}
\label{subsection:gemini-appendix}
\paragraph{Trajectory Representation}

For behavioral planning evaluation, models predict a sequence of normalized 2D image-space trajectory points conditioned on the task instruction and the initial scene image. The first predicted point is constrained to lie on the selected tool, while the remaining points describe the predicted manipulation trajectory. 

To simplify evaluation across models with different output styles, trajectories are represented as sparse waypoint sequences rather than dense per-frame motion predictions. Predicted trajectories are converted back to pixel coordinates using the image resolution before evaluation. We prompted the model to generate 32 points, enough to create a dense table-top action task trajectory. 

\paragraph{Ground-Truth Trajectory Construction}

Ground-truth trajectories are derived from tracked tool pose sequences recovered from the object pose solver. For each frame, the tool translation component is projected into image space using calibrated camera intrinsics, producing a dense 2D tool trajectory. The projected trajectory is then arc-length resampled to match the number of predicted points generated by the model.

The first ground truth trajectory is generated by taking the center point of the tool mesh and applying transforms to that throughout. We additionally evaluate a correspondence-aware trajectory variant that anchors evaluation to the predicted interaction point. Specifically, we identify the tool mesh vertex whose first-frame projection is nearest to the predicted starting point, and transport this vertex through the full pose sequence before projection into image space. If the predicted initial point does not lie directly on the tool surface, we associate it with the nearest projected mesh vertex in the first frame before transporting that vertex through the pose sequence. This produces a more interaction-centric trajectory target with more fair comparison than using only the tool-center translation, as we observed during preliminary experiments that Gemini Robotics ER does not always generate a consistent starting point, sometimes initializing trajectories near the tool center, grasp region, or intended contact point depending on the task and tool geometry.

\paragraph{Qualitative Results}
\begin{figure*}[t]
\centering

\begin{subfigure}{0.32\textwidth}
    \includegraphics[width=\linewidth]{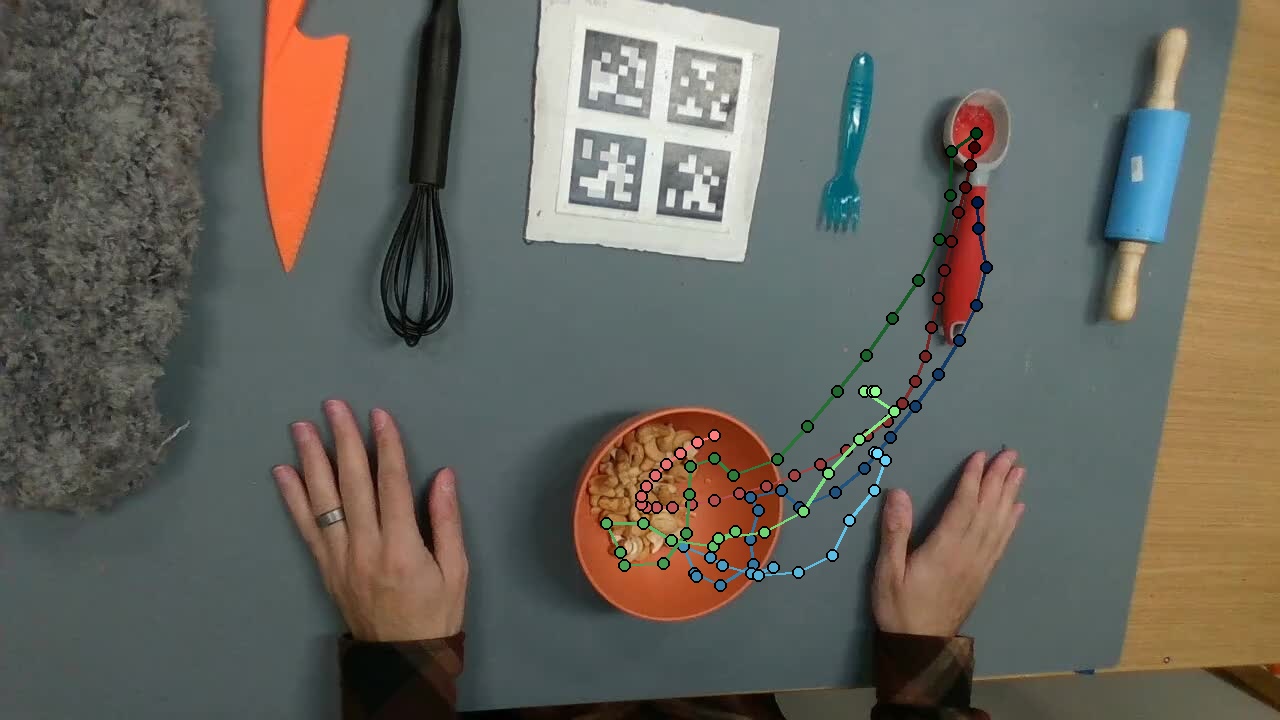}
    \caption{Scooper scooping nuts.}
\end{subfigure}
\hfill
\begin{subfigure}{0.32\textwidth}
    \includegraphics[width=\linewidth]{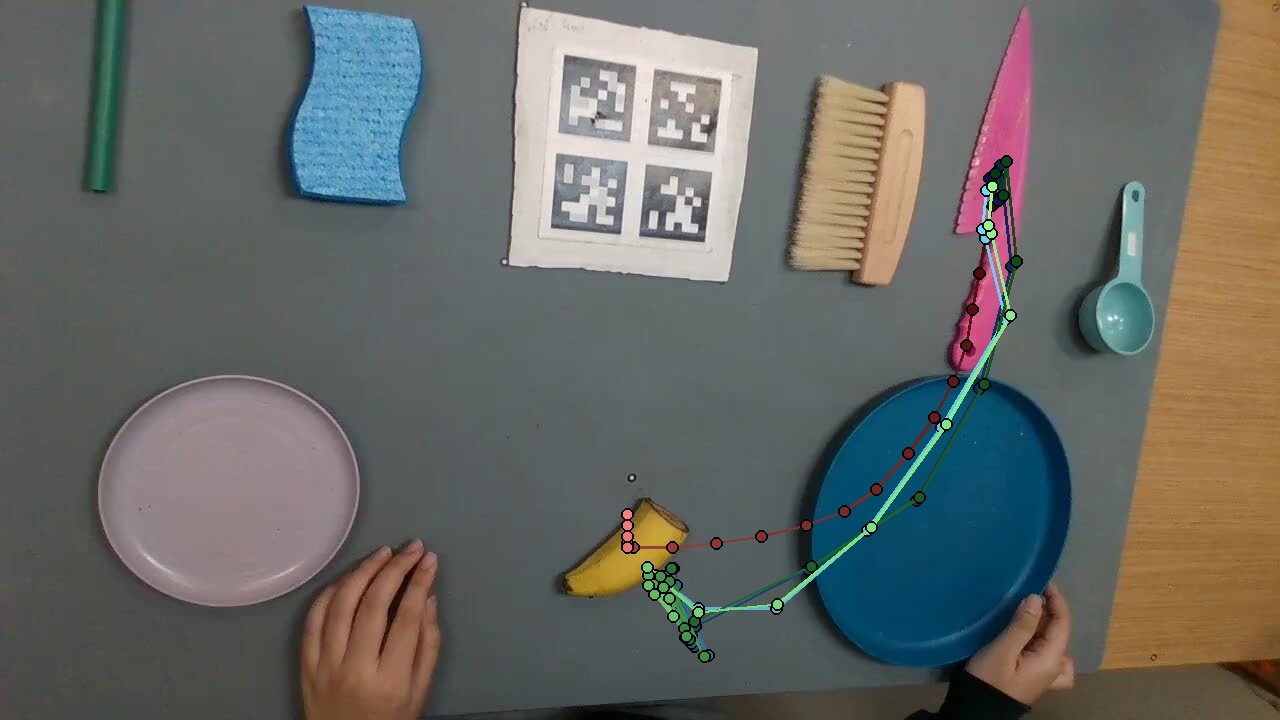}
    \caption{Knife slicing banana.}
\end{subfigure}
\hfill
\begin{subfigure}{0.32\textwidth}
    \includegraphics[width=\linewidth]{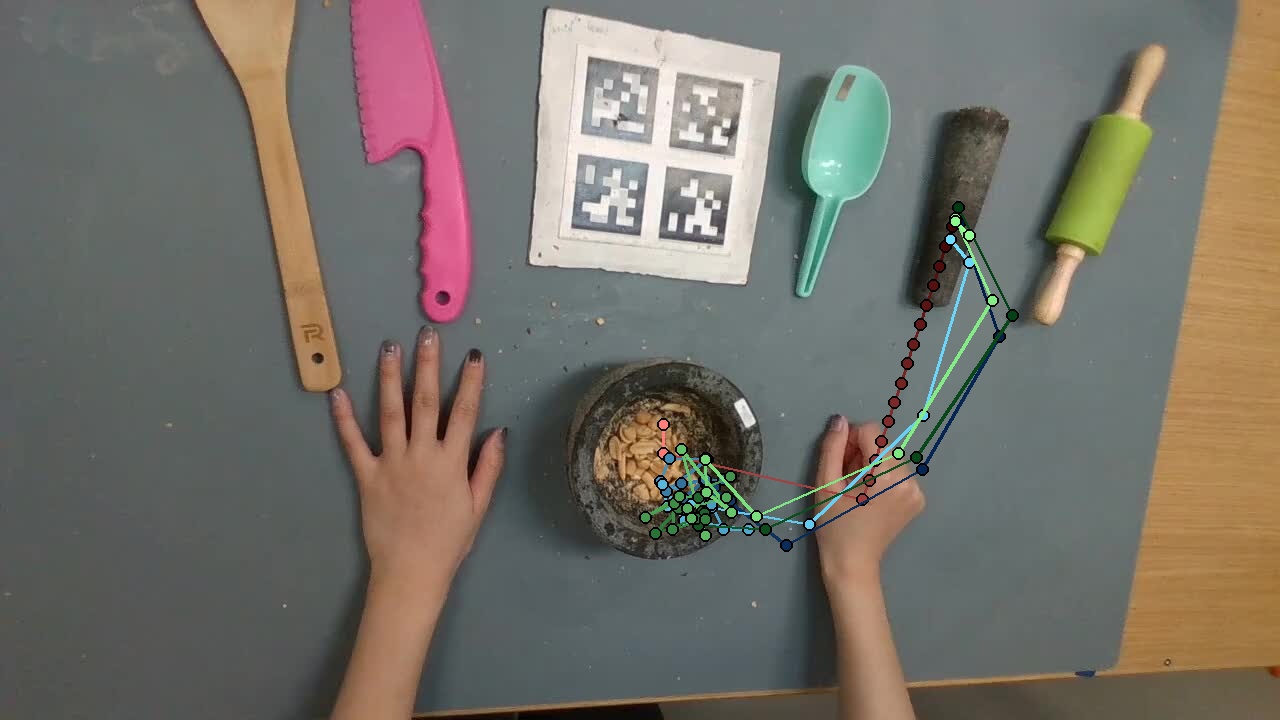}
    \caption{Pestle crushing nuts.}
\end{subfigure}

\vspace{0.5em}

\begin{subfigure}{0.32\textwidth}
    \includegraphics[width=\linewidth]{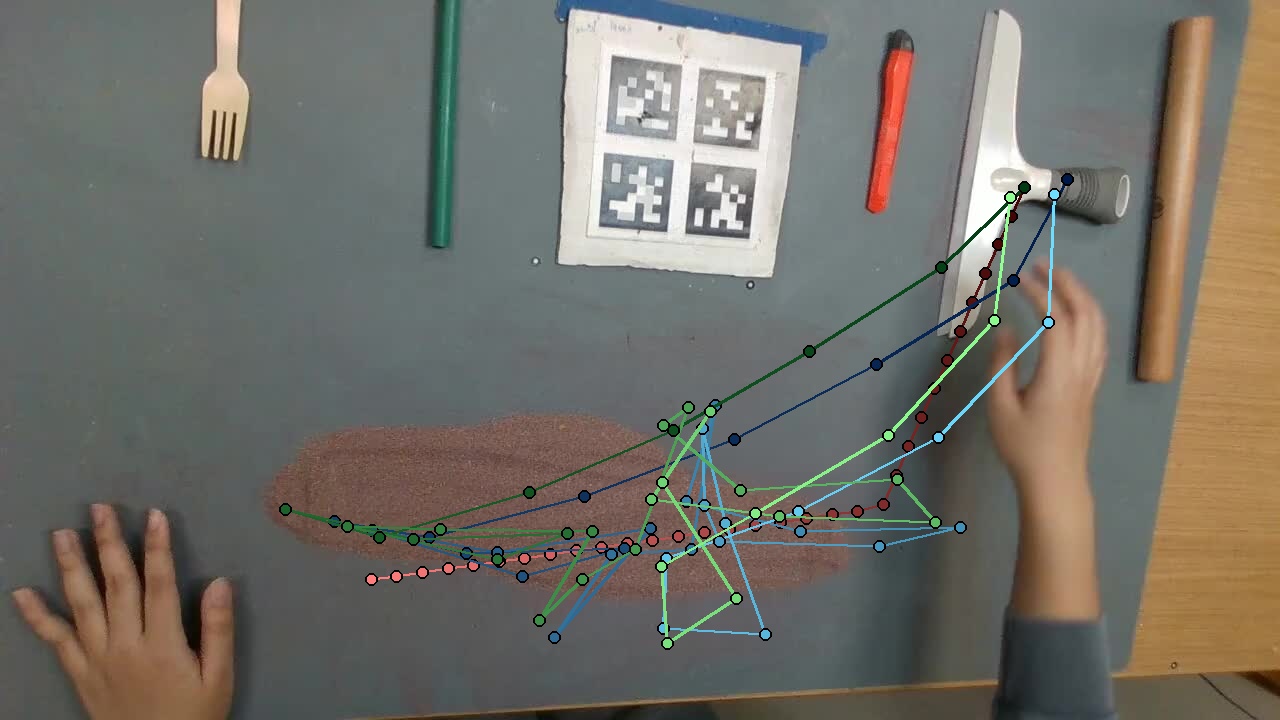}
    \caption{Squeegee collecting sand.}
\end{subfigure}
\hfill
\begin{subfigure}{0.32\textwidth}
    \includegraphics[width=\linewidth]{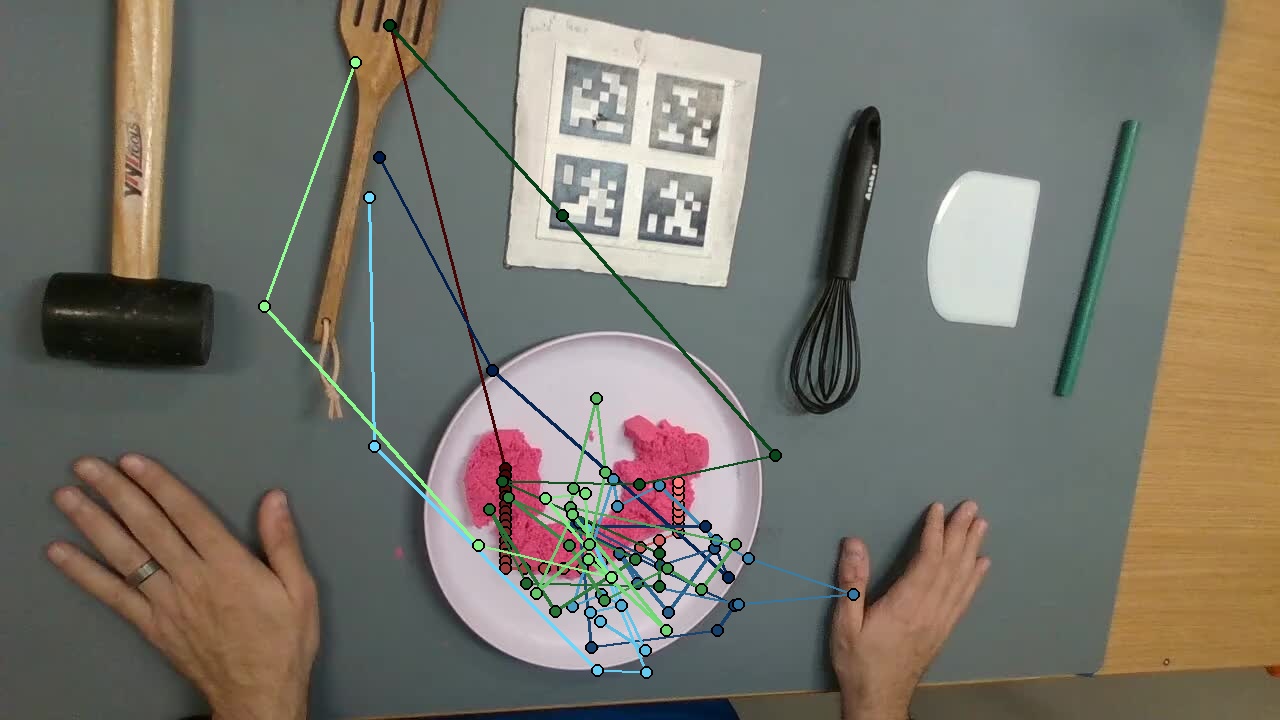}
    \caption{Fork shaping kinetic sand.}
\end{subfigure}
\hfill
\begin{subfigure}{0.32\textwidth}
    \includegraphics[width=\linewidth]{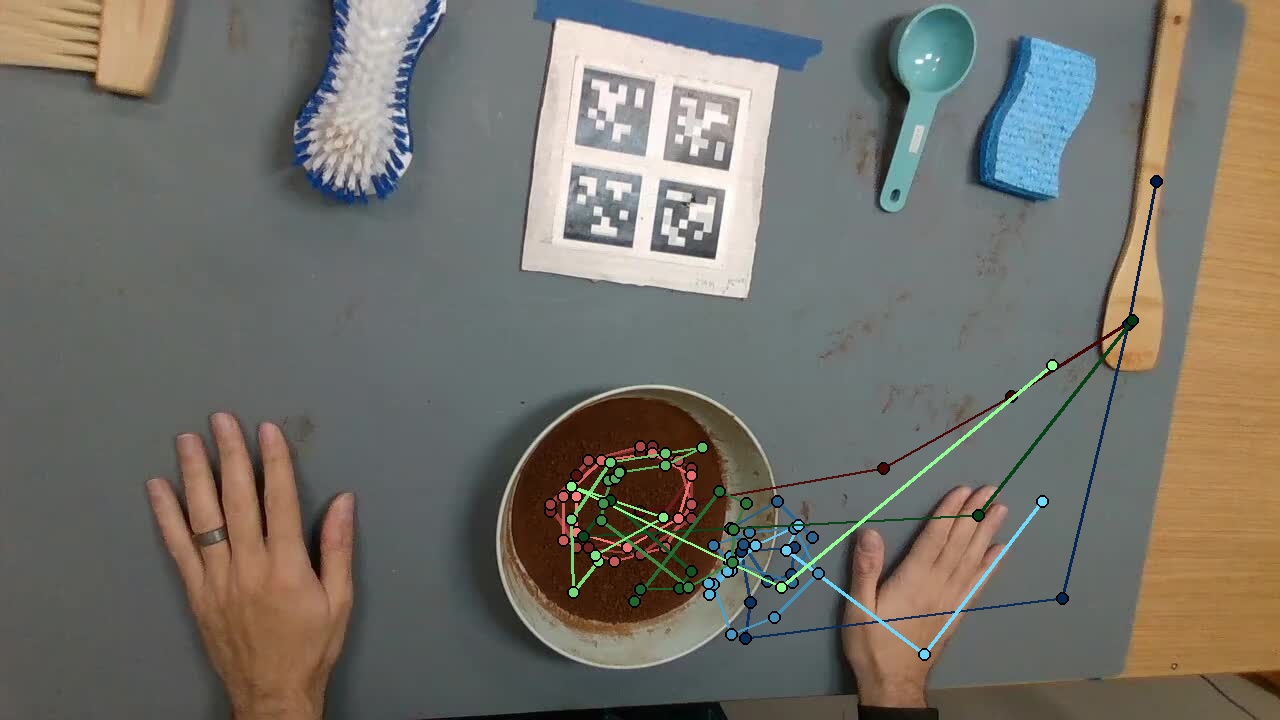}
    \caption{Stirring coffee.}
\end{subfigure}

\caption{
Qualitative trajectory generation results from Gemini Robotics ER across six manipulation scenarios. Red trajectories denote the predicted trajectories generated by Gemini Robotics ER, blue trajectories correspond to the projected tool-center ground-truth trajectories, and green trajectories correspond to the interaction-point transported trajectories. Predicted trajectories generally capture the high-level interaction geometry and intended tool-use behavior, particularly for short-horizon manipulation tasks.
}
\label{robotics-er-res}
\end{figure*}

It can be seen in \autoref{robotics-er-res} that tasks with relatively well-defined motions—typically those involving short actions, clear end states, or trajectories that start directly on the target object—tend to achieve stronger results. For example, the banana slicing trajectory(b) closely matches the ground-truth cutting motion, despite only requiring the model to infer the cup position and cutting direction from the scene context.

However, for more complex tasks, such as shaping kinetic sand into a circular structure (e), the model often oversimplifies the interaction dynamics and fails to fully capture the physical difficulty of the task. In these scenarios, successful manipulation requires more than simply tracing a circular trajectory; the tool must repeatedly reposition, push material forward, and gradually accumulate shape over multiple passes. The generated trajectories instead tend to approximate only the high-level geometric pattern of the motion.

Nevertheless, these results still demonstrate the strong capability of the model to capture the overall structure and intent of tool-use trajectories across a wide range of diverse manipulation settings.

\subsection{Policy Learning Pipeline Details}
\label{appendix:policy_pipeline}

This section provides additional implementation details for the policy
learning baselines described in the main text, including retargeting,
training, and evaluation.

\subsubsection{Retargeting}
\label{appendix:retargeting}

The evaluated policies use retargeted robot trajectories derived from the
human hand and tool motion annotations. Different retargeting procedures
are used according to the robot embodiment and policy formulation.

\paragraph{ManipTrans + Inspire.}
For ManipTrans~\cite{li2025maniptransefficientdexterousbimanual}, we follow
its hand retargeting procedure and optimize Inspire hand joint angles to
match the MANO reference, with emphasis on fingertip alignment. Since this
setting controls only the dexterous hand, the resulting joint-space
trajectory is used as the reference for residual policy learning.

\paragraph{Dex-X.}
For Dex-X~\cite{chen2026dexxlearningvisualtactiledexterous}, the human
demonstration is retargeted to the Franka FR3 + Sharpa hand embodiment.
The retargeting jointly maps wrist motion and hand articulation into the
robot arm--hand configuration space while respecting robot joint limits and
reachable configurations. We follow the retargeting and simulation pipeline
described in Dex-X and use the resulting trajectories as references for
interaction-aware reinforcement learning.

\begin{figure}[h]
\centering
\includegraphics[width=0.7\linewidth]{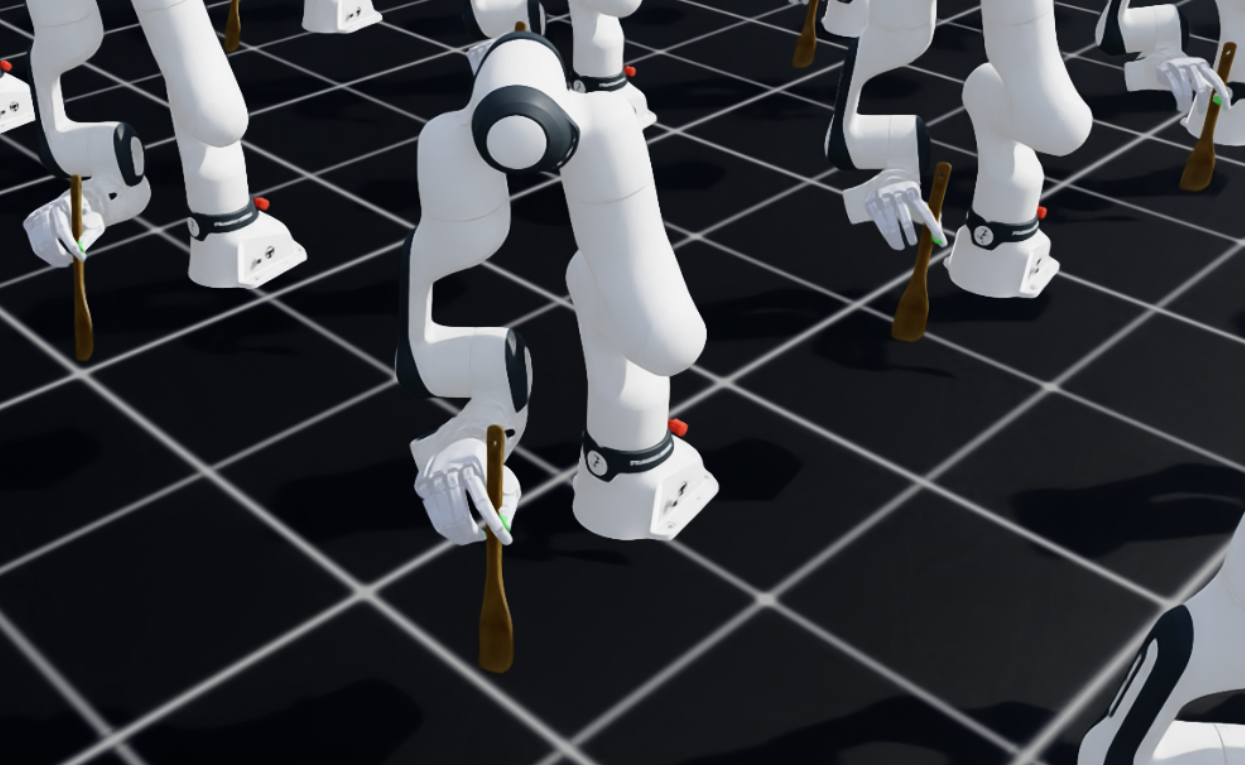}
\caption{Visualization of the tool retargeting.}
\label{fig:retarget_pipeline}
\end{figure}

\subsubsection{Evaluation Protocol}
\label{appendix:eval_protocol}

We evaluate six task signatures spanning four action categories:
stir, scoop, crush, and collect. Demonstrations contain 254--600 frames
($\approx$4--10,s at 60,Hz). Each method is evaluated over 1{,}000
episodes per task with randomized initial states and reference start
frames.

Following the reference-tracking evaluation protocol, an episode is
considered successful if the policy reaches the end of the demonstrated
interaction without violating the task-specific tracking and stability
constraints. The same success definition is used consistently across
methods whenever applicable.

\subsubsection{Training Configuration}
\label{appendix:training_config}

For ManipTrans(RL), we use the training configuration of
ManipTrans~\cite{li2025maniptransefficientdexterousbimanual}. A residual policy is trained with PPO in IsaacGym for each task using the same hyperparameter configuration across the evaluated task set.

For Dex-X (RL), we follow the released training pipeline of
Dex-X~\cite{chen2026dexxlearningvisualtactiledexterous}. Policies are trained with reinforcement learning in IsaacLab using the retargeted human demonstrations as interaction references. We use the same training pipeline across all evaluated tasks without task-specific policy design.

\subsection{Limitations and future directions.} \label{sec:limitations}
Despite its scale and diversity, \dataname~has several limitations. 
First, the dataset focuses primarily on tabletop setting in indoor environments, and therefore does not cover the full range of human tool-use behaviors found in large-scale outdoor or mobile settings. 
Second, while we include rigid, granular, and deformable target objects, the physical coverage of materials and interaction dynamics remains incomplete. 
Third, although the ego-exo capture setup enables high-fidelity reconstruction and tracking, some annotations may still contain noise due to occlusion or challenging contact configurations.  
In addition, the VLM benchmarking pipeline is intentionally permissive to account for multiple plausible tool choices. While this better reflects real-world tool use, future work could incorporate human-validated sets of acceptable tools.

\end{document}